\documentclass[final,5p,times,twocolumn,authoryear]{elsarticle}

\usepackage{float}
\usepackage{booktabs}
\usepackage{array}
\usepackage{makecell}
\usepackage{tabularx}
\usepackage{xcolor}
\usepackage{amssymb}
\usepackage{amsmath}
\usepackage{lipsum}
\usepackage{xurl}
\usepackage{hyperref}

\journal{High Energy Astrophysics}

\begin{document}

\begin{frontmatter}



\title{Multimodal Violence Understanding and Temporal Retrieval in Movies with Adaptive Audio-Visual Fusion}


\author[inst1,inst2]{Jiaqian Li} 
\author[inst1]{Rohitash Chandra} 
\affiliation[inst1]{
    organization={Transitional Artificial Intelligence Research Group, 
    School of Mathematics and Statistics, UNSW Sydney},
    city={Sydney},
    country={Australia}
}

\affiliation[inst2]{
    organization={Centre for Artificial Intelligence and Innovation, Pingla Institute},
    city={Sydney},
    country={Australia}
}

\begin{abstract}


Violence in films and streaming media remains a concern for audience protection and creates substantial workload for manual content classification. Existing violence-detection methods often rely on limited modalities or short video clips, making it difficult to capture complementary evidence and localise violent events in long, untrimmed videos. In this study, we develop a framework for multimodal video understanding that classifies violence severity, predicts multiple violence types, and locates violent events in time. We combine temporally aligned visual and audio features extracted using pretrained encoders through attention-based adaptive fusion, followed by Transformer-based temporal modelling. Our framework supports three complementary tasks: video-level violence-severity classification, multi-label violence-type prediction, and temporal localisation of violent events. We evaluate the framework on the XD-Violence benchmark and examine its cross-domain performance using AIRTLab. On XD-Violence, our multimodal model achieves 93.27\% severity-classification accuracy and a micro-F1 score of 88.60\% for multi-label violence-type prediction, outperforming the corresponding unimodal configurations. We achieve a mean temporal intersection-over-union (tIoU) of 0.5075 for temporal localisation. Our cross-domain experiments reveal performance degradation under different recording conditions, while our movie case studies demonstrate the framework's potential to retrieve potentially violent scenes for further review. Our results highlight the practical value of multimodal intelligent systems in supporting human reviewers with the identification, classification, and temporal retrieval of violent content in films and streaming media.

\end{abstract}




\end{frontmatter}




\section{Introduction}
\label{sec:introduction}

Violence has been a prominent feature of popular cinema, with content analyses documenting an increase in its prevalence among commercially successful films, including those accessible to younger audiences. Bushman et al. \citep{bushman2013gun} found that overall violence in a sample of top-selling films more than doubled between 1950 and 2012, while gun violence in films rated PG-13 more than tripled between 1985 and 2012. Extending these findings, Romer et al. \citep{romer2017continuing} reported that gun violence in PG-13 films continued to increase through 2015, indicating that this trend persisted beyond the period examined in the earlier study. More recently, Fotouhi et al. \citep{fotouhi2025trends} broadened the investigation of cinematic violence beyond visually depicted actions by examining violent language in movie dialogue between 1970 and 2020. Together, these studies highlight the importance of examining violence across different forms of cinematic expression, including physical actions and spoken language, and motivate systematic approaches to identifying and analysing violent content in films and streaming media.




Film-rating and classification systems are limited, although they can help viewers and parents judge whether a film is suitable for a particular audience to some extent. Rating systems such as the  Motion Picture Association  in the United States \citep{mpa_ratings}, the British Board of Film Classification (BBFC) Guidelines \citep{bbfc_guidelines}, and the Australian Classification \citep{australian_classification} scheme aim to help audiences make informed viewing decisions based on age suitability and content impact. \cite{australian_classification}, for example, classification decisions consider classifiable elements such as themes, violence, language, drug use, nudity, and sex, while consumer advice provides additional information about impactful and frequent content. However, these systems rely on human review, so decisions can be affected by subjective judgement, cultural expectations, and contextual interpretation. \cite{romer2014parental} showed that parents can become desensitised to violent and sexual content after repeated exposure, while \cite{romer2018parental} found that parents' judgements about gun violence in PG-13 films are influenced by whether the violence appears justified. These studies suggest that parental guidance and human classification decisions may vary according to context, exposure, and interpretation.  AI-assisted BBFC  classification illustrated that AI tools can flag potentially contentious material \citep{guardian2026bbfc}, including violence, nudity, and strong language, but human compliance officers retain responsibility for the final rating. Automatic violence detection is therefore best treated as decision support for human experts, rather than as a replacement for them.

Early work on violent-scene detection \citep{clarin2005dove, chen2011violence} focused mainly on visual cues, including blood, weapons, fighting, rapid motion, and body movement. These cues can identify visually explicit violence, but they do not cover every case. A scene may convey violence through screams, gunshots, explosions, threatening speech, abusive language, or subtitles when the visual signal is unclear, stylised, or absent. Researchers have consequently combined more than one media source of evidence. \cite{lin2009violence} combined auditory and visual classifiers through co-training and demonstrated the feasibility of detecting violent movie scenes using both modalities. Building on audio-visual analysis for movie retrieval, \cite{eyben2013affective} evaluated segment-level acoustic and visual features on 18 Hollywood films, highlighting the importance of acoustic features associated with sudden changes in sound intensity. More recent research has explored deep learning-based multimodal approaches, with \cite{pang2021violence} introducing attention-guided audio-visual fusion and reporting improved violence-detection performance on XD-Violence, while \cite{wu2020not} introduced a large audio-visual violence dataset with weak supervision. \cite{sharma2022cnn} used audio-event recognition for automated violence classification and rating in Prime Video content. Text data provides a complementary source of evidence since information from movie scripts and subtitles can reveal contextual information about violence and abuse   that may not be directly observable from visual or acoustic signals. \cite{martinez2019violence} predicted violence ratings from movie scripts, \cite{martinez2020victim} analysed violent character portrayals in script language, and \cite{na2024leveraging} combined visual, audio, and textual information. A multimodal approach can therefore capture forms of violence that any single modality may miss.

Automatic violence classification can make rating workflows more efficient by providing timestamp-level evidence. Rather than assigning one label to an entire film, a system can retrieve violent segments, identify their start and end times, estimate their severity, and summarise exposure across the video. Human experts, parents, and streaming platforms can then review the relevant scenes directly. This is useful for long-form content, where violent scenes may be brief, scattered throughout the film, or shaped by a complicated narrative context. In addition to severity, identifying the specific form of violence is important because a video may contain different or multiple violent events, such as fighting, shooting, riots, abuse, accidents, or explosions. Past studies have shown that audio-visual streams can support violence detection in Hollywood films and web videos, as well as parental-control and movie-rating applications \citep{tirupattur2021violence,pang2021violence,sharma2022cnn}. 


In our previous study, we developed a framework that combined YOLO-based frame localisation with a hybrid 3D-Convolutional Neural Network-Long Short-Term Memory (LSTM) model to classify violent segments and generate timestamped video-level violence summary  for movies \citep{chandra2026deep}. Although the framework demonstrated the feasibility of automatically locating and assessing violent scenes in movies, it relied primarily on visual information and remained sensitive to complex cinematic conditions such as lighting, camera viewpoints, occlusion, and action choreography. It also could not exploit complementary acoustic or linguistic evidence, such as gunshots, screams, threats, or abusive dialogue. Moreover, the framework primarily focused on violence severity rather than explicitly identifying multiple violence types that may coexist within the same video. These limitations motivated us to combine audio-visual evidence with temporal modelling, so that video-level assessment and localisation use the same representation.

In this study, we extend this direction by presenting a context-guided multimodal framework for violence understanding and retrieval in films and streaming media. We combine temporally aligned visual and acoustic representations through adaptive multimodal fusion, in which modality contributions are estimated from video-level context, followed by Transformer-based temporal modelling. The framework supports three complementary outputs, including video-level violence-severity classification, multi-label violence-type prediction, and timestamp-level localisation of violent events. In contrast to our previous visual-based pipeline, our framework uses complementary audio-visual evidence within a shared temporal representation, enabling a more complete characterisation of what violence occurs, its overall severity, and where it occurs in the video. The resulting outputs provide targeted evidence that can support parental guidance, content moderation, and rating workflows while reducing the manual screening burden.

\section{Related Work}

\subsection{Violence Detection}

Early violence-detection studies used handcrafted visual features to describe the appearance and motion associated with violent events. Later work placed greater emphasis on motion; \cite{hassner2012violent} introduced Violent Flows, which represents changes in optical-flow magnitude and uses a linear support vector machine to classify violent crowd behaviour. \cite{gao2016violence} extended this approach with Oriented Violent Flows by modelling the directional distribution of optical-flow vectors and applying AdaBoost-based feature selection. \cite{zhou2018violence} combined appearance and motion by detecting motion regions, extracting local histograms of oriented gradients and optical flow, and encoding them with a bag-of-words representation. These methods improved the representation of visible movement, but they rely on predefined features. They are therefore less suited to subtle, dialogue-driven, off-screen, or cinematically complex violence.


Deep learning marked a significant shift from handcrafted descriptors to automatically learned spatiotemporal representations of video data.  \cite{tran2015learning} introduced C3D, a three-dimensional convolutional neural network (3D-CNN) that uses 3D convolutional kernels to learn spatial appearance and temporal motion patterns jointly from consecutive video frames. These learned features provide a foundation for video understanding tasks, including action recognition and violent-event detection. Later studies combined convolutional features with ConvLSTM temporal modelling \citep{sudhakaran2017learning}, developed end-to-end 3D-CNN classifiers for short surveillance sequences \citep{ullah2019violence}, and used bidirectional ConvLSTMs to incorporate context from both temporal directions \citep{hanson2019bidirectional}. \cite{peixoto2021harnessing} treated violence as a heterogeneous category that can include fights, blood, explosions, and gunshots. This concept-level view is important because violence is not a single homogeneous event class; different forms of violence may exhibit distinct visual and acoustic characteristics and may occur together within the same video. For movie content, our earlier work combined YOLO-based violent-frame localisation with a lightweight 3D-CNN-LSTM model for clip- and video-level assessment \citep{chandra2026deep}. Despite these advances, visual models remain affected by lighting, camera angles, occlusion, complex choreography, and ambiguous actions. They also cannot directly detect verbal abuse, threats, or violence that occurs off-screen.

Audio-based violence detection focuses on distinctive acoustic events associated with physical danger. \cite{valenzise2007scream} showed that audio alone can detect and localise screams and gunshots when visual evidence is absent or obscured. Studies moved from individual events to movie-level analysis. \cite{eyben2013affective} evaluated segmental audio features in Hollywood films, including energy, Mel-frequency cepstral coefficients (MFCC), spectral flux, harmonicity, and sharpness. The study found that audio signals can help identify gunfire, explosions, aggressive vocalisations, and other events that are off-screen or visually unclear. However, acoustic models have limited contextual awareness. They may miss silent violence or mistake loud, non-violent events for violence.

Text-based studies examine violence through the language of scripts and dialogue. \cite{martinez2019violence} predicted expert-assigned violence ratings for 732 Hollywood scripts using lexical, semantic, sentiment, and abusive-language features. At the subtitle level, \cite{vonboguszewski2021hateful} labelled movie utterances as hateful, offensive, or normal and found that contextual BERT representations outperformed bag-of-words and Bidirectional-LSTM models.   \cite{chandra2025longitudinal} analysed subtitles from more than one thousand Hollywood films released between 1950 and 2024 and reported a gradual increase in abusive content in movie dialogue (subtitles) in recent decades, with notable variation across film genres. Recent work has also evaluated the capabilities of large language models (LLMs) in generating movie reviews, comparing models such as GPT-4o, Gemini 2.0, and DeepSeek-V3 \cite{sands2026evaluation}, highlighting the growing role of LLMs in automated movie-content analysis.  Beyond cinema, \cite{chandra2025music} examined Billboard song lyrics over seven decades and found a substantial increase in explicit and abusive language, particularly from the 1990s. These studies demonstrate that harmful content can also depend on semantic and narrative context. Violent media also raises concerns about possible effects on children and adolescents. Research has associated exposure to violent media with aggressive thoughts, desensitisation, and aggressive behaviour, although these relationships depend on individual circumstances and social context. \cite{huesmann2007impact} reviewed theoretical and empirical work suggesting that exposure to electronic media violence can increase the risk of aggressive behaviour in the short and long term, particularly among younger viewers. \cite{worth2008exposure} found that adolescents may still watch highly violent films despite age-based restrictions. \cite{coker2015media} reported an association between media violence exposure and physical aggression among fifth-grade children. In a study, \cite{dillon2017effects} found that children who watched gun violence in films showed greater interest in real guns, which makes violent media exposure a concern, particularly for younger and more vulnerable viewers.



\subsection{Multimodal Violence Detection}

Early multimodal systems combined handcrafted audio and visual descriptors; \cite{nam1998audiovisual} combined acoustic cues with colour and motion features to describe violent scenes in films. \cite{lin2010violence} used separate visual and auditory representations, including visual localisation of violent regions. \cite{perperis2011multimodal} combined detected screams, gunshots, fights, and explosions through ontology-based reasoning. Benchmark studies then evaluated handcrafted audio-visual systems for localising violent scenes in professionally produced films \citep{demarty2014benchmarking}. These approaches combined several useful signals, but they still depended on predefined descriptors and domain-specific rules.

Deep learning shifted multimodal violence detection from handcrafted feature combination to learned cross-modal interaction. \cite{wu2020not} introduced the XD-Violence dataset, a large-scale benchmark containing long, untrimmed audio-visual videos with weak video-level labels, together with a Holistic and Localised Network (HL-Net) for weakly supervised violence detection. HL-Net learns snippet-level violence scores by modelling long-range and local relationships among temporal snippets, demonstrating how complementary visual and audio information can be used in untrimmed videos. Later methods used mutual audio-visual attention, bilinear interactions, and dependency modelling so that one modality could inform the interpretation of the other \cite{pang2021violence,pang2023audiovisual}. In these models, visual features help identify actions and objects, while audio can supply evidence of gunshots, screams, and explosions that may be unclear or absent in the image stream.

Later work expanded the visual branch beyond visual frames by incorporating optical flow and semantic structure. Optical-flow-aware fusion explicitly modelled motion intensity alongside visual frames and audio, while semantic local-to-global embedding decomposed violence into concepts such as fights, blood, screams, and explosions and reduced the heterogeneity between modalities \citep{xie2022optical,pu2022semantic}. More recent studies recognised that visual frames, motion, and audio may differ in temporal scale, representation space, and reliability, motivating consistency learning, hyperbolic representation learning, and explicit feature alignment before fusion \citep{peng2024hyperbolic,jin2025aligning}. 

Although previous studies have explored multi-class violence classification and semantic representations of violence \citep{perperis2011multimodal,tirupattur2021violence,peixoto2021harnessing}, the explicit video-level prediction of multiple co-occurring violence types remains an important research direction. In contrast to approaches that focus primarily on binary detection or temporal localisation, our approach jointly predicts whether a video contains violence and identifies the different forms of violence depicted, enabling a more fine-grained characterisation of violent content.

\subsection{Temporal Localisation of Violent Scenes}


For retrieval applications, detecting violence is only the first step; systems must also identify when violent events occur within a video. Early violence-detection approaches primarily classified short, pre-segmented clips, assuming that event boundaries were already known. Therefore, weakly supervised methods extended detection to long, untrimmed videos by dividing them into temporal snippets and learning snippet-level anomaly scores from video-level labels. \cite{sultani2018real} established this Multiple-Instance Learning (MIL) formulation for anomaly detection in surveillance videos, treating each anomalous video as a bag containing at least one anomalous snippet. In the violence-detection domain, \cite{wu2020not} introduced XD-Violence, a large-scale audio-visual benchmark, and proposed HL-Net to estimate snippet-level violence scores under weak video-level supervision. HL-Net enables the approximate localisation of violent segments in untrimmed videos by modelling both local and long-range temporal relationships, without requiring precise temporal annotations. Subsequent approaches have sought to improve localisation accuracy through techniques, such as pseudo-label refinement and robust feature-magnitude learning, reducing reliance on densely annotated training data \citep{feng2021mist,tian2021weakly,wu2023weakly}.

Natural-language video localisation provides another route to temporal retrieval by predicting boundaries that match a textual query. VSLNet treats moment retrieval as span prediction and estimates the start and end of the matching video segment \citep{zhang2020span}. Moment-DETR uses direct set prediction to produce relevant intervals and query-dependent saliency scores on QVHighlights \citep{lei2021detecting}. Language-aware anomaly methods also use textual prompts to improve weakly supervised localisation \citep{yang2024text}. These studies inform a violence-specific retrieval framework that first predicts whether a full video contains violence and its overall severity, then returns timestamped violent intervals for videos classified as violent.

\section{Methodology}
\label{sec:methodology}

\subsection{Dataset and Problem Formulation}
\label{sec:dataset_problem}

\begin{figure*}[htbp!]
\centering
\includegraphics[width=\textwidth]{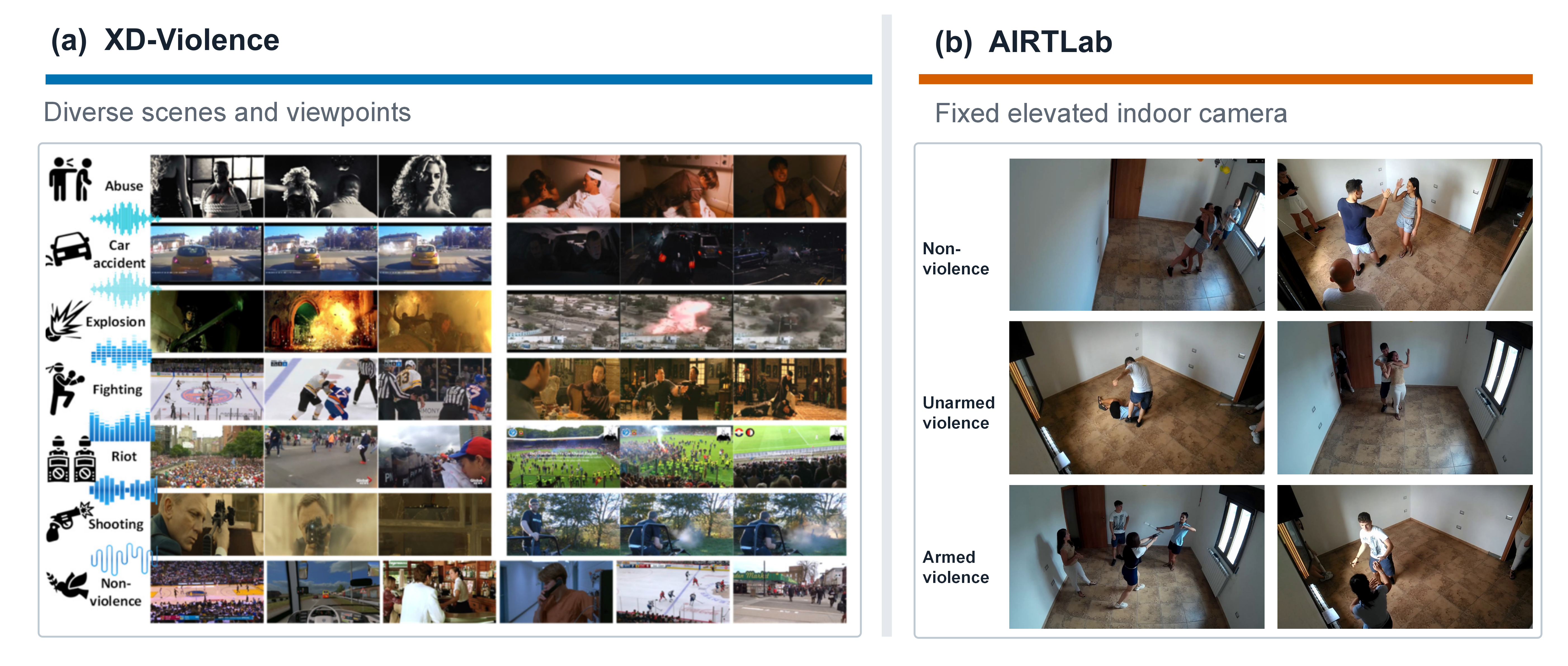}
\caption{Representative visual comparison of XD-Violence and AIRTLab. (a) XD-Violence contains heterogeneous videos spanning multiple violence categories, scene environments, and camera viewpoints \citep{wu2020not}. (b) AIRTLab consists of staged human actions recorded from fixed, elevated camera viewpoints in a comparatively controlled indoor environment \citep{bianculli2020dataset}. The contrast in scene composition, viewpoint, and recording conditions illustrates the domain shift considered in the cross-domain evaluation.}
\label{fig:dataset_comparison}
\end{figure*}


We use XD-Violence as the primary dataset and AIRTLab for cross-domain evaluation. Figure~\ref{fig:dataset_comparison} provides a visual comparison of the two datasets and highlights the differences in scene composition, camera viewpoint, and recording conditions. XD-Violence contains 4,754 untrimmed videos with approximately 217 hours of audiovisual content, covering six violence categories: abuse, fighting, car accidents, explosions, shootings, and riots, together with normal activities \citep{wu2020not}. The dataset contains diverse sources, including movie scenes, web videos, and surveillance-style footage. In particular, its inclusion of a large amount of movie content and both visual and audio information makes it suitable for the multimodal movie-violence setting considered in this study. The official split contains 3,954 training videos with video-level labels and 800 test videos, for which the start and end frames of violent events are temporally annotated \citep{wu2020not}. These temporal annotations also enable evaluation of violence localisation. The six native violence categories are additionally retained as video-level violence-type labels, allowing the framework to identify the specific forms of violence present in a video rather than only their overall severity.

\begin{table*}[h]
\centering
\caption{Operational violence-severity mapping used for XD-Violence and AIRTLab.}
\label{tab:severity_mapping}
\small
\begin{tabular}{lll}
\toprule
\textbf{Level} & \textbf{XD-Violence} & \textbf{AIRTLab} \\
\midrule
0: Non-violence & Normal & Non-violent actions \\
1: Low violence & Abuse, fighting & Pushing, slapping, fighting, kicking, punching \\
2: High violence & Car accident, explosion, shooting, riot & Shooting, stabbing, clubbing \\
\bottomrule
\end{tabular}
\end{table*}

Neither dataset provides independent ordinal severity ratings. We derive operational three-level severity targets from their annotated categories, following the AIRTLab grouping in our previous work \cite{chandra2026deep}, as shown in Table~\ref{tab:severity_mapping}.

For XD-Violence videos containing multiple violence types, the highest mapped level is assigned; a video labelled with both fighting and shooting is therefore high violence. This mapping leaves the original six type labels intact for multi-label prediction.

\subsection{Multimodal Feature Representation}
\label{sec:multimodal_features}

For each video of duration $D$, we uniformly sample $N=32$ temporal anchors over the complete video:
\begin{equation}
t_i=\frac{iD}{N}, \qquad i=0,\ldots,N-1.
\label{eq:temporal_sampling}
\end{equation}
At each anchor, we select the nearest RGB frame and a centred $\pm1$~s audio window. Audio at a video boundary is padded on the missing side so that the window remains centred. A frozen CLIP ViT-L/14-336 image encoder \cite{radford2021clip} produces a 768-dimensional visual vector for each frame. These vectors are $L_2$-normalised. A frozen BEATs encoder \cite{chen2023beats} processes each audio window. Its internal temporal embeddings are mean-pooled to produce one 768-dimensional audio vector per anchor. The resulting 32 audio vectors remain separate and are later attention-pooled to estimate the audio modality weight. The visual and audio sequences are aligned by their shared anchor indices, without an additional synchronisation step.

\begin{figure*}[h]
    \centering
    \includegraphics[width=1\linewidth]{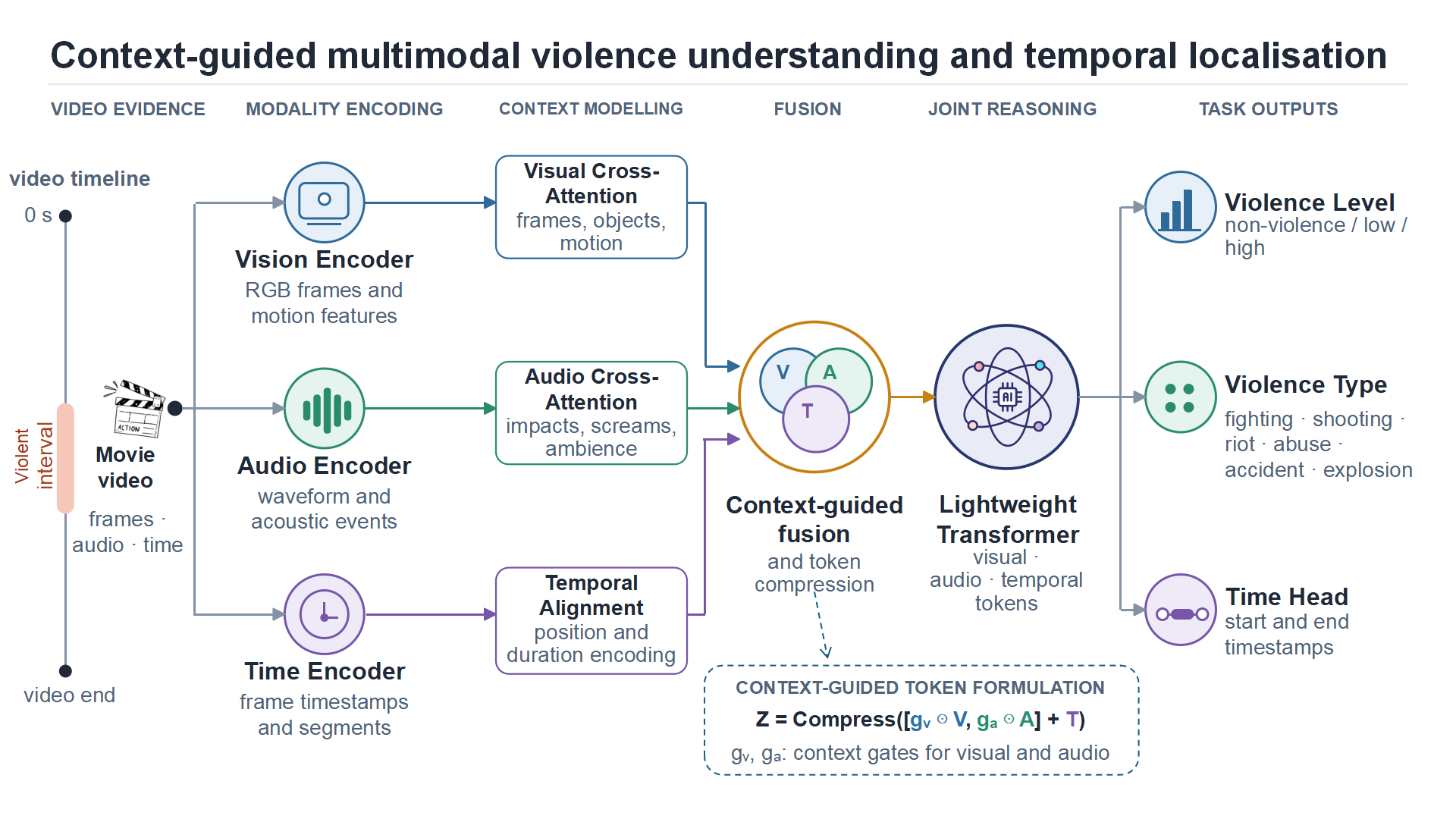}
    \caption{Overview of the proposed context-guided multimodal framework for video-based violence understanding and temporal localisation. Visual, audio, and temporal information are encoded from uniformly sampled video timestamps, modelled through modality-specific context modules, and combined by adaptive multimodal fusion. A lightweight Transformer captures temporal dependencies in the fused sequence, from which task-specific heads predict the video-level violence severity, one or more violence types, and the start and end timestamps of violent events.}
    \label{fig:violence_multimodal_llm_framework}
\end{figure*}

\subsection{Framework for Context-Guided Multimodal Violence Model}
\label{sec:context_guided_model}

Our framework addresses three tasks: video-level classification into non-violence, low violence, or high violence; video-level multi-label prediction of the six XD-Violence types; and timestamp-level localisation of violent intervals. Type labels are independent, so a video can contain several forms of violence, while a non-violent video has no positive type label. Violence-type prediction is evaluated on XD-Violence, while AIRTLab is used for cross-domain evaluation of severity. The localisation output identifies intervals for human review.

Figure~\ref{fig:violence_multimodal_llm_framework} illustrates the proposed multimodal violence framework given the temporally aligned visual and audio representations described in Section~\ref{sec:multimodal_features}.We begin with Stage 1: \textit{Video Evidence Extraction}, where we uniformly sample each input video at 32 temporal anchors. At each anchor, we extract the nearest RGB frame and a centred audio window of approximately two seconds. This approach provides temporal coverage while reducing computational costs, although brief events occurring between sampled anchors may be missed.

We proceed to Stage 2: \textit{Modality Encoding}, where we independently process the sampled frames and audio windows using the frozen CLIP ViT-L/14-336 and BEATs encoders, respectively. Each encoder produces a 768-dimensional feature vector representing visual content or acoustic events, such as fighting, gunshots, explosions, and screams. We align the resulting feature sequences using their shared temporal-anchor indices. In Stage 3: \textit{Context Modelling and Temporal Alignment}, we project the visual and audio features into a shared 256-dimensional latent space using separate projection layers. We then apply modality-specific attention pooling to identify informative observations and generate video-level summaries. We also incorporate timestamp embeddings to preserve temporal information for subsequent sequence modelling.

We continue with Stage 4: \textit{Adaptive Multimodal Fusion}, where we use a trainable gating network to estimate the relative contributions of visual and audio information from their video-level summaries. We apply the resulting modality weights to the corresponding features at each temporal anchor to construct a fused representation. This approach allows us to adapt the balance between modalities to the video content while preserving its temporal structure. In Stage 5: \textit{Temporal Reasoning}, we process the fused features, augmented with timestamp embeddings, using a lightweight, two-layer Transformer encoder. Its self-attention mechanism captures dependencies across temporal anchors, enabling us to interpret individual observations in the context of the wider video. We use these contextualised representations to support both video-level classification and temporal localisation.

Finally, in Stage 6: \textit{Task-Specific Prediction}, we use three output heads to generate complementary predictions. The severity head classifies each video as non-violent, low-violence, or high-violence. The violence-type head uses six independent sigmoid outputs to identify fighting, shooting, riot, abuse, car accident, and explosion, allowing multiple labels per video. The localisation head estimates violence probabilities at individual temporal anchors. We group consecutive high-scoring anchors into intervals and convert them into timestamps. Together, these outputs enable us to assess violence severity, identify violence types, and retrieve potentially violent scenes for further review.


\subsection{Technical details}

We separately project the aligned CLIP and BEATs vectors  into a shared $H=256$ dimensional space, yielding visual tokens $\mathbf{v}_i^{h}$ and audio tokens $\mathbf{a}_i^{h}$ at anchor $i\in\{0,\ldots,N-1\}$. For each modality $m\in\{v,a\}$, let $\mathbf{x}_{v,i}^{h}=\mathbf{v}_i^{h}$ and $\mathbf{x}_{a,i}^{h}=\mathbf{a}_i^{h}$. Attention pooling identifies the anchors that contribute most to a video-level summary:
\begin{equation}
\begin{aligned}
e_{m,i}&=\mathbf{w}_m^{\top}\tanh\!\left(W_m^{\mathrm{att}}\mathbf{x}_{m,i}^{h}+\mathbf{b}_m^{\mathrm{att}}\right),\\
\beta_{m,i}&=\frac{\exp(e_{m,i})}{\sum_{j=0}^{N-1}\exp(e_{m,j})},\\
\mathbf{r}_m&=\sum_{i=0}^{N-1}\beta_{m,i}\mathbf{x}_{m,i}^{h},\qquad m\in\{v,a\}.
\end{aligned}
\label{eq:attention_pooling}
\end{equation}
where  $e_{m,i}$ is the attention score, $\beta_{m,i}$ is its weight after normalisation over the $N$ anchors, and $\mathbf{r}_m$ is the resulting modality summary. The model learns the parameters $W_m^{\mathrm{att}}\in\mathbb{R}^{128\times H}$, $\mathbf{b}_m^{\mathrm{att}}\in\mathbb{R}^{128}$, and $\mathbf{w}_m\in\mathbb{R}^{128}$ separately for visual and audio tokens.

We then use the two summaries to determine video-level modality weights through a trainable gating network $f_g$:
\begin{equation}
\begin{aligned}
[g_v,g_a]&=\operatorname{MaskedSoftmax}\!\left(f_g([\mathbf{r}_v;\mathbf{r}_a])\right),\\
\mathbf{c}_i&=2\left(g_v\mathbf{v}_i^{h}+g_a\mathbf{a}_i^{h}\right).
\end{aligned}
\label{eq:adaptive_fusion}
\end{equation}
The brackets denote concatenation, $g_v$ and $g_a$ are the visual and audio weights, and $\mathbf{c}_i$ is the fused token at anchor $i$. The masked softmax excludes an unavailable modality; when both are present, $g_v+g_a=1$. The factor of two compensates for the scale reduction caused by weighting two available modalities whose weights sum to one. One pair of weights is computed per video and applied to every anchor, retaining the full temporal sequence. We then add the projected timestamp embeddings  to the fused tokens, and a two-layer Transformer encoder \citep{vaswani2017attention} models dependencies across the video.

We share the Transformer output by three task-specific heads, where mean-pooled temporal tokens feed the three-class severity head and the six-label violence-type head. The severity head uses softmax, while the type head uses independent sigmoid outputs to allow multiple types per video. A sigmoid localisation head operates on each temporal token, assigning a violence probability to its corresponding anchor. We provide details for the architectural settings  in Appendix~\ref{app:implementation_details}.

\subsection{Training and Inference}
\label{sec:training_inference}

We keep the  CLIP and BEATs encoders frozen during training, and the trainable part of the framework consists of the modality projections, temporal-attention pooling, adaptive fusion module, Transformer encoder, and task-specific output heads. This design allows the pretrained encoders to provide general visual and acoustic representations, while the proposed model learns violence-specific multimodal and temporal relationships.

We train the video-level severity model with categorical cross-entropy, while the six independent type labels use binary cross-entropy. For localisation, each sampled anchor receives a binary target according to whether it falls inside an annotated violent interval; binary cross-entropy supervises the token scores. Video-level severity and type experiments optimise their respective losses using the shared backbone and separate heads. When violence-type prediction and temporal localisation are trained jointly, the objective is
\begin{equation}
\mathcal{L}
=
\lambda_{\mathrm{type}}\mathcal{L}_{\mathrm{type}}
+\lambda_{\mathrm{loc}}\mathcal{L}_{\mathrm{loc}}
+\lambda_{\mathrm{sev}}\mathcal{L}_{\mathrm{sev}}^{\mathrm{derived}}
+\lambda_{\mathrm{cons}}\mathcal{L}_{\mathrm{cons}},
\label{eq:joint_loss}
\end{equation}
where  $\mathcal{L}_{\mathrm{type}}$ and $\mathcal{L}_{\mathrm{loc}}$ are binary cross-entropy terms. The auxiliary $\mathcal{L}_{\mathrm{sev}}^{\mathrm{derived}}$ is the negative log-likelihood of the mapped severity target computed from type probabilities. The consistency term $\mathcal{L}_{\mathrm{cons}}$ aligns the maximum temporal violence score with a detached violence-presence probability derived from the type predictions. Localisation loss applies to videos with temporal annotations and to confirmed non-violent videos. The loss weights are given in ~\ref{app:implementation_details}.

At inference, the severity label is the class with the highest softmax probability. We threshold each type probability  independently using a value selected on validation data; hence zero, or more than one types may be returned. We group the localisation scores above the fixed threshold $\tau_{\mathrm{loc}}$  into contiguous intervals and converted back to video timestamps, and provide the threshold settings  in ~\ref{app:implementation_details}.

\begin{table}[h]
\centering
\caption{Research questions and corresponding experimental evaluations.}
\label{tab:research_questions}
\footnotesize
\begin{tabular}{p{0.04\columnwidth} p{0.38\columnwidth} p{0.42\columnwidth}}
\hline
\textbf{RQ} & \textbf{Research question} & \textbf{Evaluation} \\
\hline
RQ1 &
Does combining visual and acoustic information improve video-level violence severity and type prediction? &
XD-Violence; CLIP-only, BEATs-only, and CLIP+BEATs comparisons for severity and type prediction. \\

RQ2 &
Can the model localise violent moments and their temporal boundaries? &
XD-Violence temporal annotations; mean tIoU and Recall@tIoU at multiple thresholds. \\

RQ3 &
How well does the model generalise across video domains? &
Cross-domain evaluation from XD-Violence to the surveillance-oriented AIRTLab dataset. \\

RQ4 &
How does the framework behave on representative real-movie and television excerpts? &
Qualitative comparison of visual-only, audio-only, and multimodal severity predictions, violence types, and retrieved temporal intervals. \\
\hline
\end{tabular}
\end{table}

For all quantitative comparisons, the same dataset partitions and evaluation protocol are maintained across competing configurations. Source-level leakage between training, validation, and test sets is avoided where applicable. Model selection, early stopping, and violence-type and localisation decision thresholds are determined using validation data only, while the final test set is reserved for reporting. Experiments are repeated across multiple random seeds, with mean and standard deviation reported for the principal metrics.

\subsection{Experiment Setup}
\label{sec:experimental_setup}
We use XD-Violence as the primary benchmark for model development and in-domain evaluation \cite{wu2020not}.We draw the training and validation data from XD-Violence and reserve a separate held-out partition for final in-domain testing. We evaluate both three-class violence-severity classification and six-label violence-type prediction on XD-Violence. We use the validation set for model selection, early stopping, and selecting the violence-type and localisation decision thresholds, while reserving the test set exclusively for final evaluation. We keep AIRTLab separate from model development and use it as an external cross-domain dataset to assess generalisation to surveillance-oriented videos \cite{bianculli2020dataset}. Since the AIRTLab action taxonomy does not directly correspond to the six XD-Violence violence types, we evaluate cross-domain performance on AIRTLab only under the three-level violence-severity formulation.

We report accuracy, macro F1, and class-wise F1 scores for non-violence, low violence, and high violence for three-class violence-severity classification.  We report macro F1, mean average precision (mAP), micro F1, exact-match accuracy, and Hamming loss for multi-label violence-type prediction. We additionally report per-type F1 scores for fighting, shooting, riot, abuse, car accident, and explosion to assess performance across individual violence categories. We evaluate temporal localisation using mean temporal intersection-over-union (tIoU) and Recall@tIoU at thresholds of [0.3, 0.5, and 0.7].

We repeat each quantitative experiment five times using five distinct random seeds to account for stochastic variation during optimisation, with one independent run per seed. We fix the seeds to 40, 41, 42, 43, and 44 for all classification experiments. We use the same seeds and data partitions across competing configurations and report the principal results as the mean $\pm$ sample standard deviation.

\section{Results}

\subsection{Classification Results}
\label{sec:classification_results}

\subsubsection{Violence-Severity Classification}
\label{sec:severity_results}

Table~\ref{tab:severity_results} compares the visual-only, audio-only, and multimodal configurations for three-level violence-severity classification. CLIP+BEATs achieves 93.27\% accuracy and a macro-F1 of 0.9208, compared with 89.97\% accuracy for BEATs and 88.75\% for CLIP. BEATs outperforms CLIP across all three severity classes, suggesting that acoustic cues provide useful evidence for distinguishing violence levels. Fusion further improves F1 for non-violence, low violence, and high violence, respectively. Low violence remains the most difficult class, likely reflecting overlap between threatening actions, physical confrontations, and more severe violence.  We find that multimodal fusion consistently achieves the strongest performance across both overall and class-wise metrics. This improvement holds for non-violence, low violence, and high violence, indicating that multimodal learning benefits are not limited to a particular severity category.

\begin{table*}[h]
\centering
\caption{Video-level violence-severity classification for   the XD-Violence dataset. We report the   mean $\pm$  standard deviation over five runs using random seeds 40--44. The best result for each metric is shown in bold.}
\label{tab:severity_results}
\small
\resizebox{\textwidth}{!}{
\begin{tabular}{llccccc}
\hline
\textbf{Modality} &
\textbf{Encoder} &
\textbf{Accuracy} &
\textbf{Macro F1} &
\textbf{Non-violence F1} &
\textbf{Low-violence F1} &
\textbf{High-violence F1} \\
\hline
Visual only &
CLIP &
$0.8875 \pm 0.0113$ &
$0.8725 \pm 0.0136$ &
$0.8561 \pm 0.0165$ &
$0.8445 \pm 0.0303$ &
$0.9170 \pm 0.0064$ \\
Audio only &
BEATs &
$0.8997 \pm 0.0080$ &
$0.8824 \pm 0.0129$ &
$0.8653 \pm 0.0272$ &
$0.8465 \pm 0.0146$ &
$0.9353 \pm 0.0033$ \\
Visual + Audio &
CLIP + BEATs &
$\mathbf{0.9327 \pm 0.0086}$ &
$\mathbf{0.9208 \pm 0.0097}$ &
$\mathbf{0.9104 \pm 0.0172}$ &
$\mathbf{0.8958 \pm 0.0127}$ &
$\mathbf{0.9563 \pm 0.0084}$ \\
\hline
\end{tabular}
}
\end{table*}

\begin{figure}[h]
    \centering
    \includegraphics[width=\columnwidth]{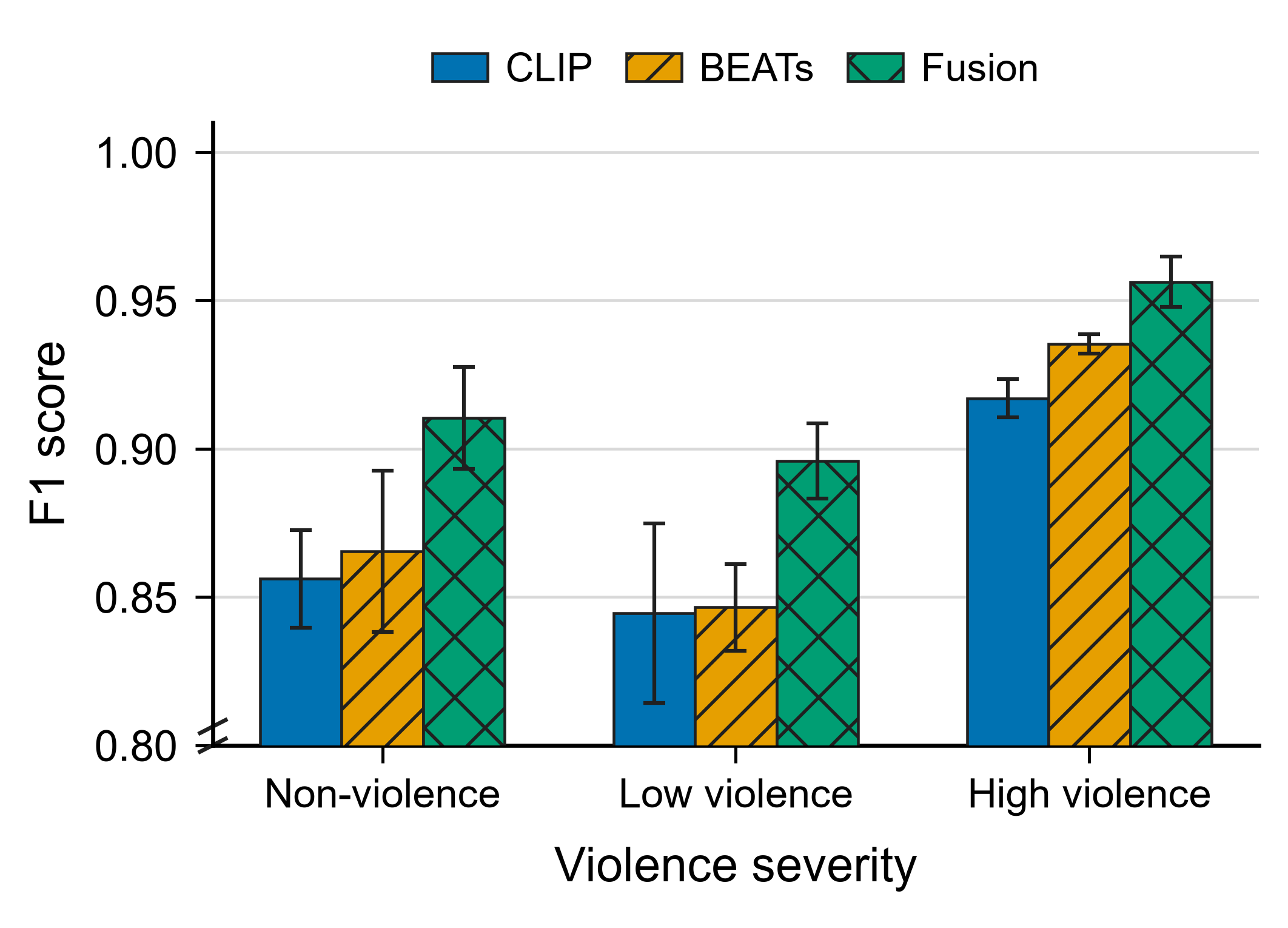}
    \caption{Class-wise F1 scores for violence-severity classification on XD-Violence. Bars show the mean across five random seeds (40--44), and error bars indicate the sample standard deviation. Fusion denotes CLIP+BEATs.}
    \label{fig:fig_exp2_per_severity_f1}
\end{figure}

Figure~\ref{fig:fig_exp2_per_severity_f1} provides a direct comparison of the class-wise F1 scores across the three modality configurations, making the contribution of visual, acoustic, and fused representations at each violence-severity level more explicit.

The consistent improvement suggests that visual and acoustic information provide complementary evidence for assessing violence severity. Visual features capture actions, objects, and scene context, whereas the audio stream can provide additional evidence through screams, impacts, gunshots, explosions, and other acoustically salient events \citep{valenzise2007scream,eyben2013affective}. Such signals may remain informative when the corresponding visual event is ambiguous, partially occluded, or not captured clearly by the sampled frame. The fused representation can therefore combine evidence that may be incomplete in either modality alone.

As illustrated in Figure~\ref{fig:fig_exp2_per_severity_f1}, the audio-only BEATs configuration consistently outperforms the visual-only CLIP configuration across all three severity levels. This indicates that acoustic information provides useful evidence not only for highly violent content, but also for distinguishing non-violent and low-violence videos. The advantage is most pronounced for high violence, which may be partly explained by the presence of acoustically salient events such as gunshots, explosions, crashes, impacts, and intense vocalisations. The soundtrack thus contributes severity information beyond what is visible in sampled frames, supporting the inclusion of BEATs.

\begin{figure}[h]
    \centering
    \includegraphics[width=\columnwidth]{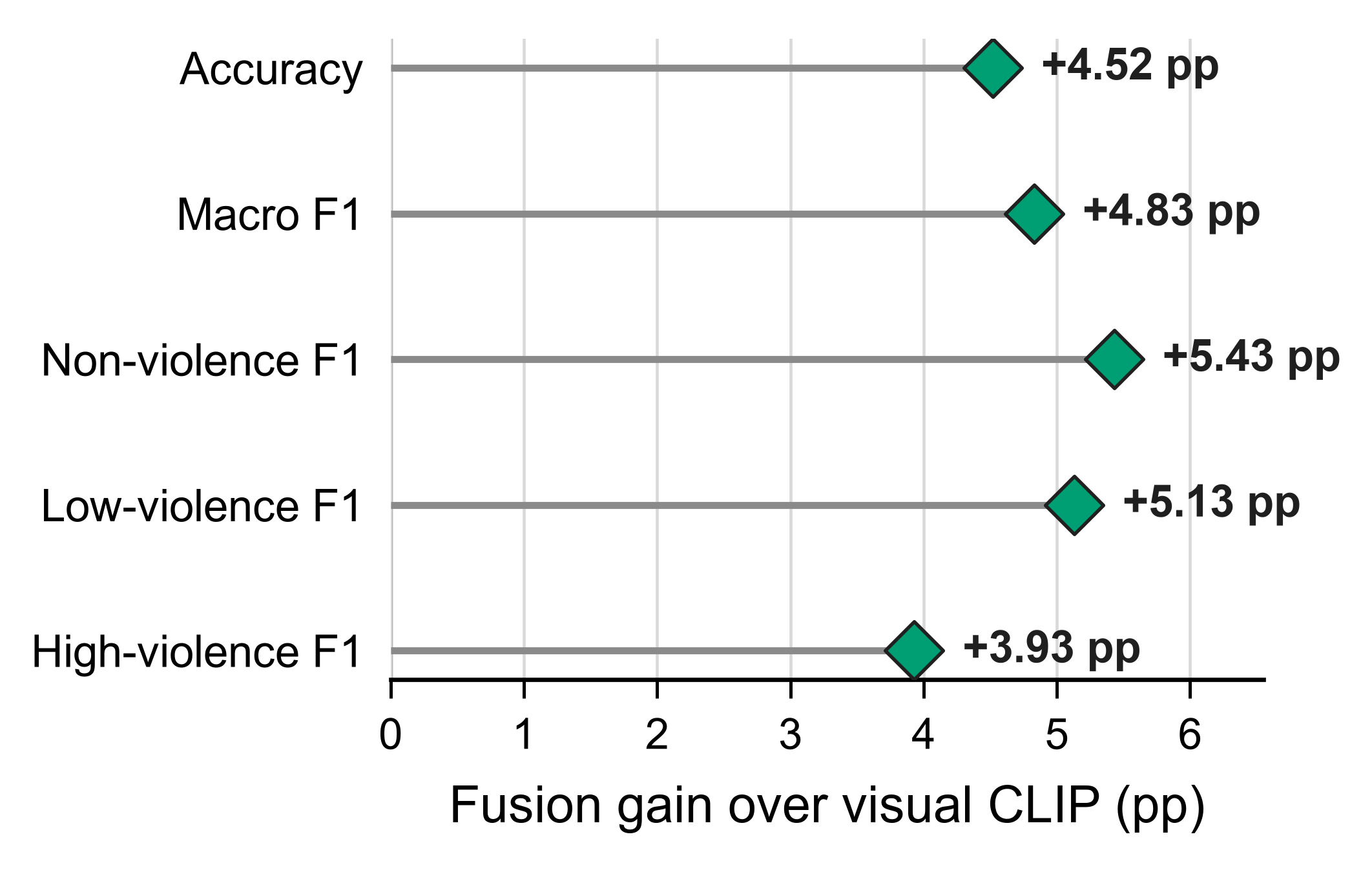}
    \caption{Absolute improvement in violence-severity classification from multimodal fusion relative to the visual-only CLIP configuration. Each point represents the difference between the mean performance of CLIP+BEATs and CLIP across five random seeds, expressed in percentage points.}
    \label{fig:fig_exp2_fusion_gain}
\end{figure}

A key observation found from Table~\ref{tab:severity_results} and Figure~\ref{fig:fig_exp2_per_severity_f1} is that low violence remains the weakest of the three severity classes. This suggests that additional modalities can reduce the ambiguity between severity levels, while contextual overlap between low- and high-violence behaviours remains an important challenge. This finding is particularly relevant in comparison with our previous visual-based framework \citep{chandra2026deep}, where distinguishing low- and high-violence content was identified as a recurring difficulty because high-violence videos may also contain actions characteristic of low violence. Figure~\ref{fig:fig_exp2_fusion_gain} illustrates the improvement obtained when acoustic information is added to the visual CLIP representation. The gain is positive for every reported severity metric, reinforcing the observation that audio provides complementary information rather than benefiting only one particular class.

\subsubsection{Violence-Type Prediction}
\label{sec:type_results}

A different modality pattern emerges for violence-type prediction. As shown in Table~\ref{tab:type_results}, visual-only CLIP provides a stronger unimodal representation than BEATs across the overall multi-label metrics, while the multimodal model still achieves the strongest overall performance. In these experiments, type prediction relies more heavily on visual semantics than severity classification does. Multimodal fusion achieves the best overall results, with a macro-F1 of 0.7805, micro-F1 of 0.8860, and mAP of 0.8378. It also reduces Hamming loss from 0.0440 to 0.0330 compared with CLIP alone.

\begin{table*}[h]
\centering
\caption{Multi-label violence-type prediction results on XD-Violence. Results are reported as mean $\pm$ sample standard deviation over five runs using random seeds 40--44. Higher values are better except for Hamming loss, for which lower values are better. The best result for each metric is shown in bold.}
\label{tab:type_results}
\small
\resizebox{\textwidth}{!}{
\begin{tabular}{llccccc}
\hline
\textbf{Modality} &
\textbf{Encoder} &
\textbf{Macro F1 $\uparrow$} &
\textbf{mAP $\uparrow$} &
\textbf{Micro F1 $\uparrow$} &
\textbf{Exact Match $\uparrow$} &
\textbf{Hamming Loss $\downarrow$} \\
\hline
Visual only &
CLIP &
$0.7682 \pm 0.0197$ &
$0.8301 \pm 0.0203$ &
$0.8483 \pm 0.0179$ &
$0.7879 \pm 0.0209$ &
$0.0440 \pm 0.0054$ \\
Audio only &
BEATs &
$0.6922 \pm 0.0260$ &
$0.7809 \pm 0.0074$ &
$0.8008 \pm 0.0195$ &
$0.7441 \pm 0.0178$ &
$0.0575 \pm 0.0053$ \\
Visual + Audio &
CLIP + BEATs &
$\mathbf{0.7805 \pm 0.0362}$ &
$\mathbf{0.8378 \pm 0.0155}$ &
$\mathbf{0.8860 \pm 0.0246}$ &
$\mathbf{0.8391 \pm 0.0234}$ &
$\mathbf{0.0330 \pm 0.0066}$ \\
\hline
\end{tabular}
}
\end{table*}

The stronger visual baseline is consistent with the nature of violence-type recognition. Categories such as fighting, riots, and car accidents are characterised by combinations of human actions, objects, participants, and scene structure that are directly represented in the visual stream. Audio alone can indicate that an intense event is occurring, but may provide less information about the exact semantic category of that event.

Nevertheless, multimodal fusion improves the overall multi-label prediction. The improvement is particularly evident in micro F1, exact-match accuracy, and Hamming loss, whereas the gain in macro F1 is comparatively modest. This pattern suggests that fusion improves the consistency of the complete predicted label set rather than uniformly increasing the performance of every individual violence type. In other words, acoustic information can complement the visual representation by correcting some label-level errors even when visual information remains the stronger individual modality.

\begin{figure} [h]
    \centering
    \includegraphics[width=1\linewidth]{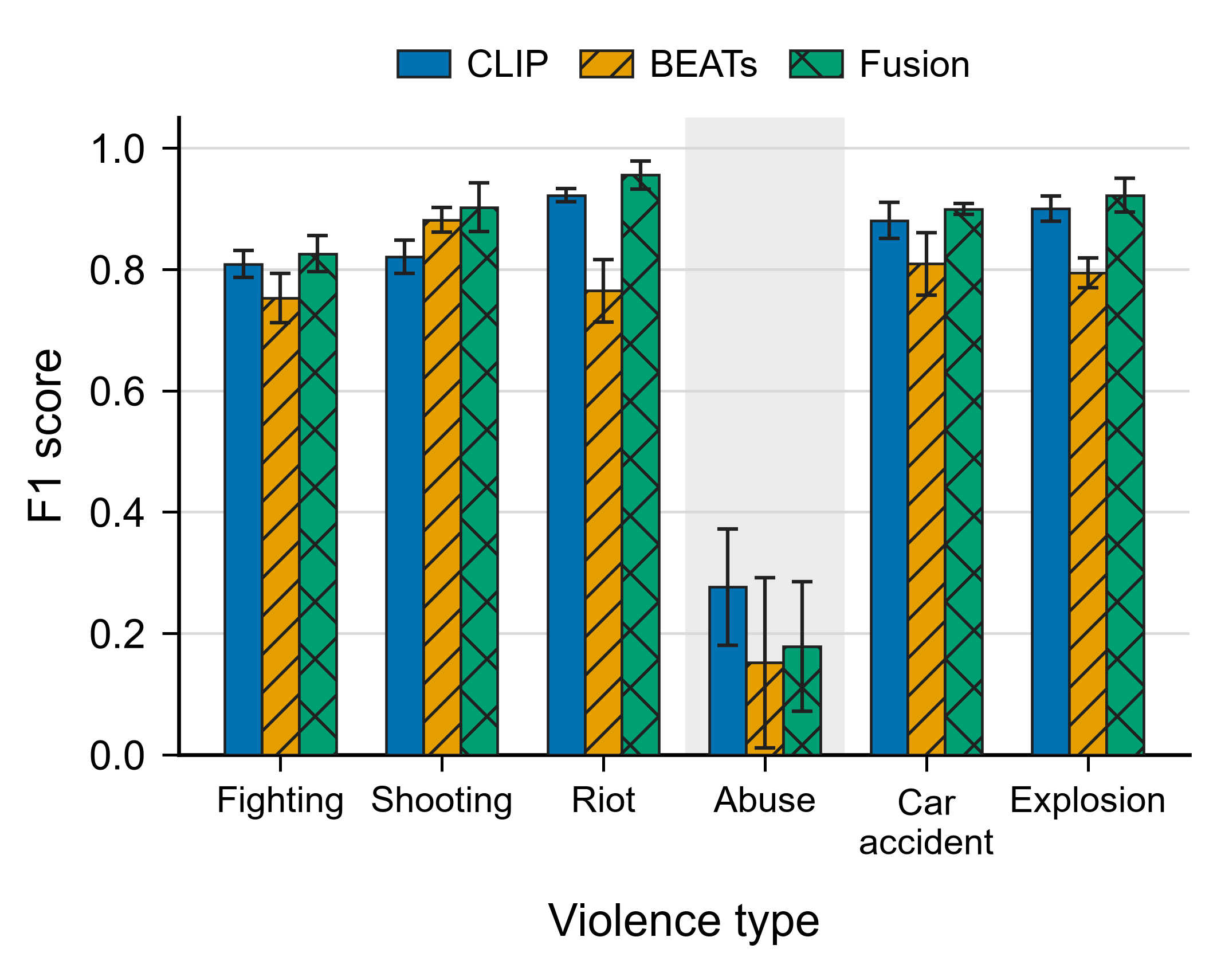}
    \caption{Per-type F1 scores for violence-type prediction on XD-Violence. Bars show the mean across five random seeds (40--44), and error bars indicate the sample standard deviation. Fusion denotes CLIP+BEATs.}
    \label{fig:fig_exp2_per_type_f1}
\end{figure}

Figure~\ref{fig:fig_exp2_per_type_f1} further shows that each modality's usefulness depends on the violence category. CLIP performs better than BEATs for most types, whereas shooting is an important exception. The relatively strong audio performance for shooting is consistent with the distinctive acoustic characteristics of gunfire reported in earlier violence-detection studies \citep{valenzise2007scream}. The fused model benefits from both sources of evidence and achieves the highest mean F1 for five of the six categories.

We find that abuse is the main exception to the general multimodal improvement. All three configurations perform substantially worse for abuse than for the other violence categories, and the larger error bars indicate greater variability across random seeds. Moreover, fusion does not improve over the visual-only configuration for this category. One possible explanation is that abuse contains more heterogeneous or subtle behaviours and may depend more strongly on contextual information than visually or acoustically distinctive events such as shootings and explosions.

These results suggest that visual features better capture the semantic characteristics of violence, while audio provides complementary evidence for categories such as shooting. Fusion achieves the highest F1 for five of the six categories. Abuse remains difficult, with lower and more variable performance across runs. This may reflect the contextual nature of abuse, which is not always identifiable from isolated visual or acoustic cues.

\subsection{Cross-Domain Generalisation}
\label{sec:cross_domain}

Table~\ref{tab:cross_domain_results} evaluates the transfer of the current CLIP-based video-level model from XD-Violence to the surveillance-oriented AIRTLab dataset. Here, cross-domain refers to training the model on XD-Violence and evaluating it directly on AIRTLab without AIRTLab-specific training, whereas in-domain refers to training using AIRTLab data and evaluating on its held-out test set. Although performance decreases under cross-domain transfer, the model retains meaningful predictive capability without being trained on AIRTLab, reaching an accuracy of $0.6057\pm0.0296$. In-domain training substantially improves performance, increasing accuracy to $0.7886\pm0.0421$ and macro F1 to $0.7921\pm0.0407$. The visual features retain some predictive value on AIRTLab, but the stronger in-domain result shows the value of adapting to the target domain.

\begin{table}[h]
\centering
\caption{Video-level violence-severity classification on AIRTLab. Cross-domain denotes training on XD-Violence followed by direct evaluation on AIRTLab, whereas in-domain denotes training and evaluation using AIRTLab. The current results are reported as mean $\pm$ sample standard deviation over five random seeds. The previous 3D-CNN-LSTM result is included as a reference from our earlier study \citep{chandra2026deep}; macro F1 was not reported at the video level in that study.}
\label{tab:cross_domain_results}
\small
\resizebox{\columnwidth}{!}{
\begin{tabular}{llcc}
\hline
\textbf{Model} &
\textbf{Setting} &
\textbf{Accuracy} &
\textbf{Macro F1} \\
\hline
CLIP &
Cross-domain &
$0.6057 \pm 0.0296$ &
$0.5834 \pm 0.0247$ \\
3DCNNLSTM &
In-domain &
$0.6615 \pm 0.0853$ &
-- \\
CLIP &
In-domain &
$\mathbf{0.7886 \pm 0.0421}$ &
$\mathbf{0.7921 \pm 0.0407}$ \\
\hline
\end{tabular}
}
\end{table}

The cross-domain result is notable because it is obtained without AIRTLab-specific training and remains within the same general performance range as the video-level result reported by our previous in-domain 3D-CNN-LSTM framework. However, the substantial improvement after training directly on AIRTLab shows that domain shift remains important. XD-Violence contains heterogeneous movie, web, and surveillance-style videos, whereas AIRTLab consists of staged actions recorded from fixed viewpoints in a controlled environment \citep{wu2020not,bianculli2020dataset}. Differences in scene structure, action presentation, camera viewpoint, and recording conditions can therefore alter the visual cues associated with violence severity. Although the model retains some cross-domain predictive ability, target-domain training substantially improves performance. Further evaluation on diverse datasets is needed to establish robust generalisation.

\subsection{Temporal Violence Localisation}
\label{sec:localisation_results}

Table~\ref{tab:localisation_results} summarises the temporal localisation performance of the multimodal CLIP+BEATs model. The model achieves a mean temporal intersection-over-union (tIoU) of $0.5075$, while recall decreases progressively as the required overlap threshold becomes stricter. This indicates that the model can identify the approximate temporal regions containing violent events, but precise recovery of their boundaries remains more difficult.

\begin{table}[h]
\centering
\caption{Temporal violence-localisation performance on XD-Violence. Results are reported as mean $\pm$ sample standard deviation across five random seeds (40--44).}
\label{tab:localisation_results}
\small
\begin{tabular}{lc}
\hline
\textbf{Metric} & \textbf{Mean $\pm$ SD} \\
\hline
Mean tIoU & $0.5075 \pm 0.0333$ \\
Recall@0.3 & $0.7210 \pm 0.0491$ \\
Recall@0.5 & $0.5099 \pm 0.0444$ \\
Recall@0.7 & $0.3236 \pm 0.0316$ \\
\hline
\end{tabular}
\end{table}

The reduction in recall from lower to higher tIoU thresholds suggests that localisation errors are primarily associated with temporal-boundary precision rather than a complete failure to identify violent regions. This is expected given that the framework represents each video using only 32 uniformly sampled temporal anchors; the resulting temporal resolution limits how precisely short or rapidly changing violent events can be aligned with their annotated start and end times. Even with this limited resolution, its approximate timestamps can direct reviewers to relevant parts of long videos.

\subsection{Modality and Qualitative Analysis}
\label{sec:qualitative_results}

As a continuation of our previous movie-oriented violence-detection study \citep{chandra2026deep}, we further examine representative movie and television scenes to investigate how the proposed multimodal framework behaves in realistic cinematic settings. Although our previous work focused primarily on visual-based violence assessment, the present analysis additionally considers modality-specific severity predictions, multimodal fusion, violence-type outputs, and retrieved temporal intervals. The tables provide a broader comparison across multiple clips, while the corresponding figures illustrate the temporal behaviour of the localisation head for selected examples.

For the qualitative case studies, the excerpts were purposively selected from the listed films and television episodes to include explicit violence as well as non-violent, threatening, and contextually ambiguous actions. One author served as the human annotator and watched each selected excerpt in full before recording a concise free-form description of the principal observed action, such as fight, club, talk, carry, run, threat, or confrontation. These descriptions were not constrained to the three severity classes or the six XD-Violence categories because their purpose was to provide an interpretable reference for what occurs in each excerpt. The annotations were produced from the video content independently of the model outputs and were subsequently used as the reference for qualitative comparison with those outputs.

\subsubsection{\textit{Game of Thrones}}
\label{sec:qual_got}

Table~\ref{tab:game_of_thrones_results} shows that the three modality configurations generally produce consistent severity predictions for the action-oriented \textit{Game of Thrones} clips, particularly when explicit fighting or weapon-related actions are present. The multimodal model also retrieves multiple violent moments within longer clips, indicating that the framework can identify separated violent events rather than assigning a single interval to the entire scene.

An interesting exception is No.~2--2. All three severity configurations classify the clip as non-violent, and the localisation head retrieves no violent interval, while the violence-type head independently predicts \emph{abuse}. In this scene, Mormont is holding Daenerys Targaryen, who is covered in blood. Although no abusive action actually occurs within the short clip, the isolated visual appearance can plausibly resemble an abuse scene when viewed without its broader narrative context. This discrepancy illustrates an important property of the multi-head framework. The severity and localisation heads may correctly indicate that no active violent event is occurring, while the type head can still respond to visual semantic cues associated with a particular form of violence. The case highlights the difficulty of interpreting short movie excerpts without sufficient narrative context, especially when blood, physical restraint, or injured characters are present.

\begin{table*}[h]
\centering
\caption{Game of Thrones: qualitative results.}
\label{tab:game_of_thrones_results}

\begingroup
\footnotesize
\setlength{\tabcolsep}{3pt}
\renewcommand{\arraystretch}{1.12}

\begin{tabular}{
    @{}
    >{\centering\arraybackslash}m{0.60cm}
    >{\centering\arraybackslash}m{1.55cm}
    >{\centering\arraybackslash}m{0.80cm}
    >{\centering\arraybackslash}m{0.80cm}
    >{\centering\arraybackslash}m{1.15cm}
    >{\raggedright\arraybackslash}m{1.55cm}
    >{\raggedright\arraybackslash}m{4.20cm}
    >{\raggedright\arraybackslash}m{2.00cm}
    @{}
}
\toprule
\textbf{No.} &
\textbf{Timeline} &
\textbf{Visual} &
\textbf{Audio} &
\makecell{\textbf{Visual}\\\textbf{+ Audio}} &
\makecell[l]{\textbf{Violence}\\\textbf{types}} &
\makecell[l]{\textbf{Retrieved moments}\\
             \textbf{Relative interval (s) [peak score]}} &
\makecell[l]{\textbf{Human}\\\textbf{annotation}} \\
\midrule

1-1 & 00:20--00:25 & low & low & low & fighting &
1.692--2.939 [0.6189] & Club \\
\cmidrule(lr){1-8}

1-2 & 01:06--01:14 & low & non & non & none &
\textemdash & Talk \\
\cmidrule(lr){1-8}

1-3 & 01:35--02:21 & low & low & low & fighting &
\makecell[l]{
18.073--36.869 [0.8150]\\
2.169--5.060 [0.7060]\\
39.760--41.206 [0.5595]} &
Club \\
\cmidrule(lr){1-8}

1-4 & 02:49--02:59 & low & low & low & fighting &
\makecell[l]{
6.719--7.656 [0.5576]\\
8.594--8.906 [0.5161]} &
Run \\
\cmidrule(lr){1-8}

1-5 & 03:20--03:28 & non & non & non & none &
\textemdash & Talk \\
\cmidrule(lr){1-8}

1-6 & 03:42--03:48 & non & non & non & none &
\textemdash & Talk \\
\cmidrule(lr){1-8}

1-7 & 04:00--04:13 & low & low & low & fighting &
\textemdash & Talk \\
\cmidrule(lr){1-8}

1-8 & 04:39--05:16 & low & low & low & fighting &
\makecell[l]{
5.231--11.044 [0.8097]\\
12.206--20.344 [0.8038]\\
27.319--28.481 [0.5516]} &
Club, Fight \\

\midrule

2-1 & 00:14--01:33 & low & low & low & fighting &
\makecell[l]{
30.742--40.580 [0.7779]\\
45.498--52.877 [0.7151]\\
11.067--13.527 [0.5218]} &
Club, Fight \\
\cmidrule(lr){1-8}

2-2 & 01:35--01:49 & non & non & non & abuse &
\textemdash & Carry \\

\bottomrule
\end{tabular}

\endgroup
\end{table*}

\begin{figure*} [h]
    \centering
    \includegraphics[width=1\linewidth]{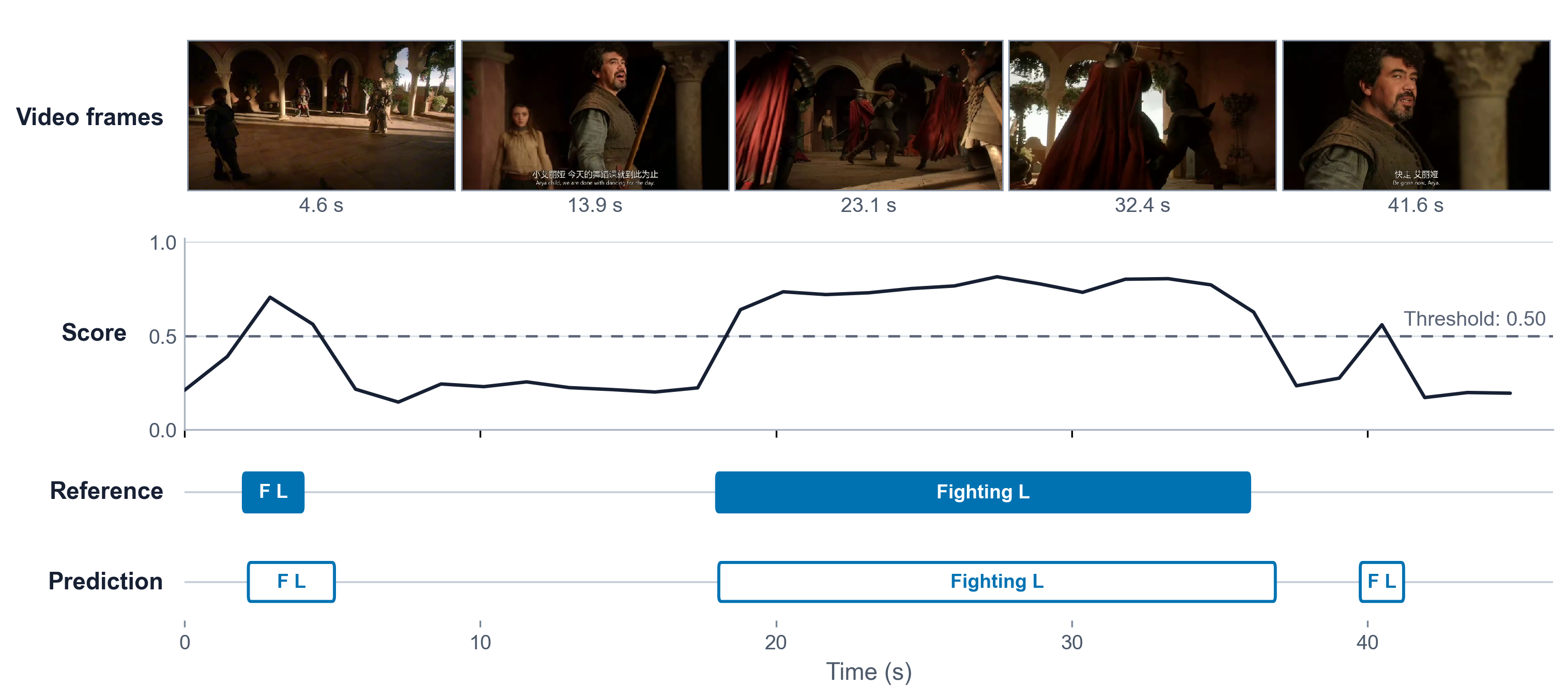}
    \caption{Qualitative temporal violence-localisation result for \textit{Game of Thrones} No.~1--3.}
    \label{fig:temporal_localisation_case_study_template_severity_type_game}
\end{figure*}

Figure~\ref{fig:temporal_localisation_case_study_template_severity_type_game} provides a detailed example for No.~1--3. The short interval at approximately 2--5~s is successfully detected even though the violent action lasts only a few seconds, during which the swordsman draws his sword and immediately attacks a soldier. The longer interval from approximately 18--36~s, where the swordsman fights a group of soldiers, is also clearly localised with a sustained violence score above the threshold. A short additional prediction occurs around 39--41~s when the character again draws his sword, although no actual fighting follows. This false positive is understandable because the motion closely resembles a preparatory fighting action and is strongly associated with imminent violence in this cinematic context.

\subsubsection{\textit{Hidden Man}}
\label{sec:qual_hidden_man}

Table~\ref{tab:hidden_man_results} reveals a different modality pattern for \textit{Hidden Man}. Several confrontation-heavy clips receive high-severity predictions from the visual-only model, whereas the audio and multimodal configurations often reduce these predictions to low or non-violent. This suggests that visual cues such as guns, threatening gestures, and aggressive body posture can overstate the apparent violence when considered alone, while acoustic information can provide complementary evidence about whether an explicit violent action is actually occurring.

No.~8 provides a representative example. In this scene, one character points a gun at another without pulling the trigger or firing a shot. The visual-only model classifies the scene as high violence, whereas the audio-only model predicts non-violence, consistent with the absence of a corresponding gunshot or other strong violent acoustic cue. After combining the two modalities, the prediction is moderated to low violence, with the violence-type head identifying fighting. This example illustrates how acoustic information can temper a strong visual violence cue without completely removing the threatening context conveyed by the scene.

Table~\ref{tab:hidden_man_results} also shows that the multimodal model retrieves the main fighting intervals in clips annotated as fights or clubbing events, although some short false-positive intervals remain. For example, No.~5 illustrates a different source of ambiguity. The clip is manually annotated as running and rolling. Visually, this motion can resemble the aftermath of a vehicle impact, where a person is thrown or rolls after being struck. This may help explain why the violence-type head predicts \emph{car accident}, while all three modality configurations classify the clip as high violence. The case highlights how visually similar body-motion patterns can lead to semantic confusion between genuine violent events and non-violent cinematic actions, particularly when only a short excerpt is considered.

\begin{table*}[h]
\centering
\caption{Hidden Man: qualitative results.}
\label{tab:hidden_man_results}

\begingroup
\footnotesize
\setlength{\tabcolsep}{3pt}
\renewcommand{\arraystretch}{1.12}

\begin{tabular}{
    @{}
    >{\centering\arraybackslash}m{0.60cm}
    >{\centering\arraybackslash}m{1.55cm}
    >{\centering\arraybackslash}m{0.80cm}
    >{\centering\arraybackslash}m{0.80cm}
    >{\centering\arraybackslash}m{1.15cm}
    >{\raggedright\arraybackslash}m{1.55cm}
    >{\raggedright\arraybackslash}m{4.20cm}
    >{\raggedright\arraybackslash}m{2.00cm}
    @{}
}
\toprule
\textbf{No.} &
\textbf{Timeline} &
\textbf{Visual} &
\textbf{Audio} &
\makecell{\textbf{Visual}\\\textbf{+ Audio}} &
\makecell[l]{\textbf{Violence}\\\textbf{types}} &
\makecell[l]{\textbf{Retrieved moments}\\
             \textbf{Relative interval (s) [peak score]}} &
\makecell[l]{\textbf{Human}\\\textbf{annotation}} \\
\midrule

1 & 00:24--00:43 & high & non & non & none &
\textemdash & Confront \\
\cmidrule(lr){1-8}

2 & 00:47--01:56 & high & low & low & fighting &
\makecell[l]{
54.825--63.425 [0.7983]\\
18.275--20.425 [0.6791]} &
Fight \\
\cmidrule(lr){1-8}

3 & 02:33--02:42 & high & low & low & fighting &
0.130--3.502 [0.7833] &
Confront \\
\cmidrule(lr){1-8}

4 & 04:10--04:44 & low & low & low & fighting &
\makecell[l]{
0.530--16.420 [0.7958]\\
19.598--21.717 [0.6315]\\
32.311--33.900 [0.5921]} &
Fight \\
\cmidrule(lr){1-8}

5 & 04:46--04:58 & high & high & high & car accident &
\makecell[l]{
2.336--3.773 [0.6506]\\
7.367--8.445 [0.5999]} &
Run, Rolling \\
\cmidrule(lr){1-8}

6 & 04:59--05:56 & low & low & low & fighting &
\makecell[l]{
40.500--57.600 [0.8552]\\
6.300--18.900 [0.7375]\\
33.300--35.100 [0.5978]} &
Fight \\
\cmidrule(lr){1-8}

7 & 06:11--06:43 & low & low & low & fighting &
\makecell[l]{
14.349--25.234 [0.8277]\\
5.443--12.370 [0.7426]\\
0.495--2.474 [0.6139]} &
Club \\
\cmidrule(lr){1-8}

8 & 07:06--07:59 & high & non & low & fighting &
19.143--20.807 [0.5150] &
Confront \\

\bottomrule
\end{tabular}

\endgroup
\end{table*}

\begin{figure*} [h]
    \centering
    \includegraphics[width=1\linewidth]{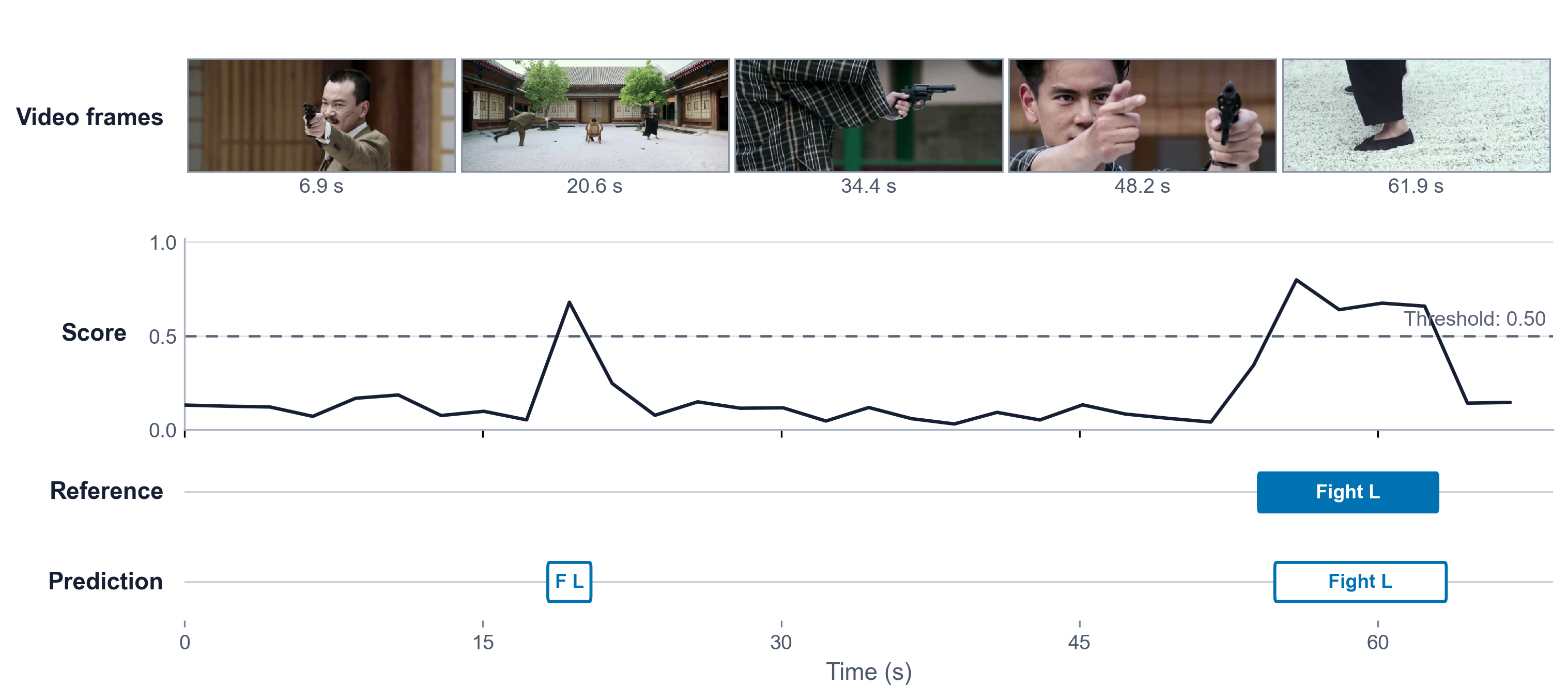}
    \caption{Qualitative temporal violence-localisation result for \textit{Hidden Man} No.~2.}
    \label{fig:temporal_localisation_case_study_template_severity_type_hidden}
\end{figure*}

Figure~\ref{fig:temporal_localisation_case_study_template_severity_type_hidden} illustrates this behaviour in No.~2. During the first half of the clip, the characters confront each other with guns or gun-like gestures, but no physical fight occurs, and the localisation score remains below the threshold for most of this period. A brief false positive appears around 18--20 s. At this point, the camera composition and background music create a strong sense of tension and suggest that violence may occur next, which may explain the elevated localisation score despite the absence of an actual fight. In contrast, the genuine fight at approximately 54--63~s is correctly localised with a sustained score above the decision threshold.

\subsubsection{\textit{Ip Man}}
\label{sec:qual_ip_man}

Table~\ref{tab:ip_man_results} shows that the model generally identifies fighting-oriented clips as low violence, while several more ambiguous scenes reveal differences between the video-level and temporal heads. No.~3 is one such example. The scene depicts a confrontation immediately before a fight, but no physical fighting has yet occurred. Nevertheless, all three severity configurations classify the clip as low violence, and the violence-type head predicts \emph{fighting}. This is understandable because the characters' body postures, spatial arrangement, and confrontational behaviour closely resemble the beginning of a fight. Interestingly, the localisation head retrieves no violent interval, suggesting that the temporal head is more conservative in identifying whether an explicit violent action is actually occurring at a particular moment in this scene.

\begin{table*}[h]
\centering
\caption{Ip Man: qualitative results.}
\label{tab:ip_man_results}

\begingroup
\footnotesize
\setlength{\tabcolsep}{3pt}
\renewcommand{\arraystretch}{1.12}

\begin{tabular}{
    @{}
    >{\centering\arraybackslash}m{0.60cm}
    >{\centering\arraybackslash}m{1.55cm}
    >{\centering\arraybackslash}m{0.80cm}
    >{\centering\arraybackslash}m{0.80cm}
    >{\centering\arraybackslash}m{1.15cm}
    >{\raggedright\arraybackslash}m{1.55cm}
    >{\raggedright\arraybackslash}m{4.20cm}
    >{\raggedright\arraybackslash}m{2.00cm}
    @{}
}
\toprule
\textbf{No.} &
\textbf{Timeline} &
\textbf{Visual} &
\textbf{Audio} &
\makecell{\textbf{Visual}\\\textbf{+ Audio}} &
\makecell[l]{\textbf{Violence}\\\textbf{types}} &
\makecell[l]{\textbf{Retrieved moments}\\
             \textbf{Relative interval (s) [peak score]}} &
\makecell[l]{\textbf{Human}\\\textbf{annotation}} \\
\midrule

1 & 00:04--01:49 & low & low & low & fighting &
\makecell[l]{
31.350--51.150 [0.7822]\\
67.650--87.450 [0.7772]\\
54.450--61.050 [0.7500]} &
Punch, Fight \\
\cmidrule(lr){1-8}

2 & 01:52--02:06 & low & low & low & fighting &
\makecell[l]{
0.000--7.150 [0.6603]\\
7.583--8.883 [0.5933]} &
Punch \\
\cmidrule(lr){1-8}

3 & 02:13--02:30 & low & low & low & fighting &
\textemdash &
Confront \\
\cmidrule(lr){1-8}

4 & 02:34--02:41 & non & non & non & none &
\textemdash &
Confront \\
\cmidrule(lr){1-8}

5 & 02:48--02:54 & non & non & non & none &
\textemdash &
Confront \\
\cmidrule(lr){1-8}

6 & 04:48--05:13 & low & low & low & fighting &
\makecell[l]{
6.773--12.352 [0.8281]\\
17.930--23.508 [0.8098]\\
1.195--4.383 [0.7231]} &
Punch, Fight \\
\cmidrule(lr){1-8}

7 & 06:39--06:51 & low & low & non & riot &
\makecell[l]{
2.672--7.659 [0.6739]\\
0.891--1.247 [0.5311]\\
1.603--2.316 [0.5098]} &
Group protest \\

\bottomrule
\end{tabular}

\endgroup
\end{table*}

No.~7 provides another informative disagreement between the task-specific heads. The violence-type head correctly identifies the group protest as \emph{riot}, whereas the visual-only and audio-only severity models classify the clip as low violence and the fused model classifies it as non-violent. Under the severity mapping used in this study, riot belongs to the high-violence category, so these severity predictions are incorrect. A possible explanation is that the observed excerpt mainly contains collective protest behaviour rather than explicit physical violence. The type head can therefore recognise the broader semantic pattern of a riot from the crowd structure and group activity, while the severity head may rely more strongly on the immediate intensity of the visual and acoustic evidence. This case highlights a distinction between recognising \emph{what kind} of event is being depicted and estimating \emph{how violent} the currently observed behaviour appears.

\begin{figure*}[h]
    \centering
    \includegraphics[width=1\linewidth]{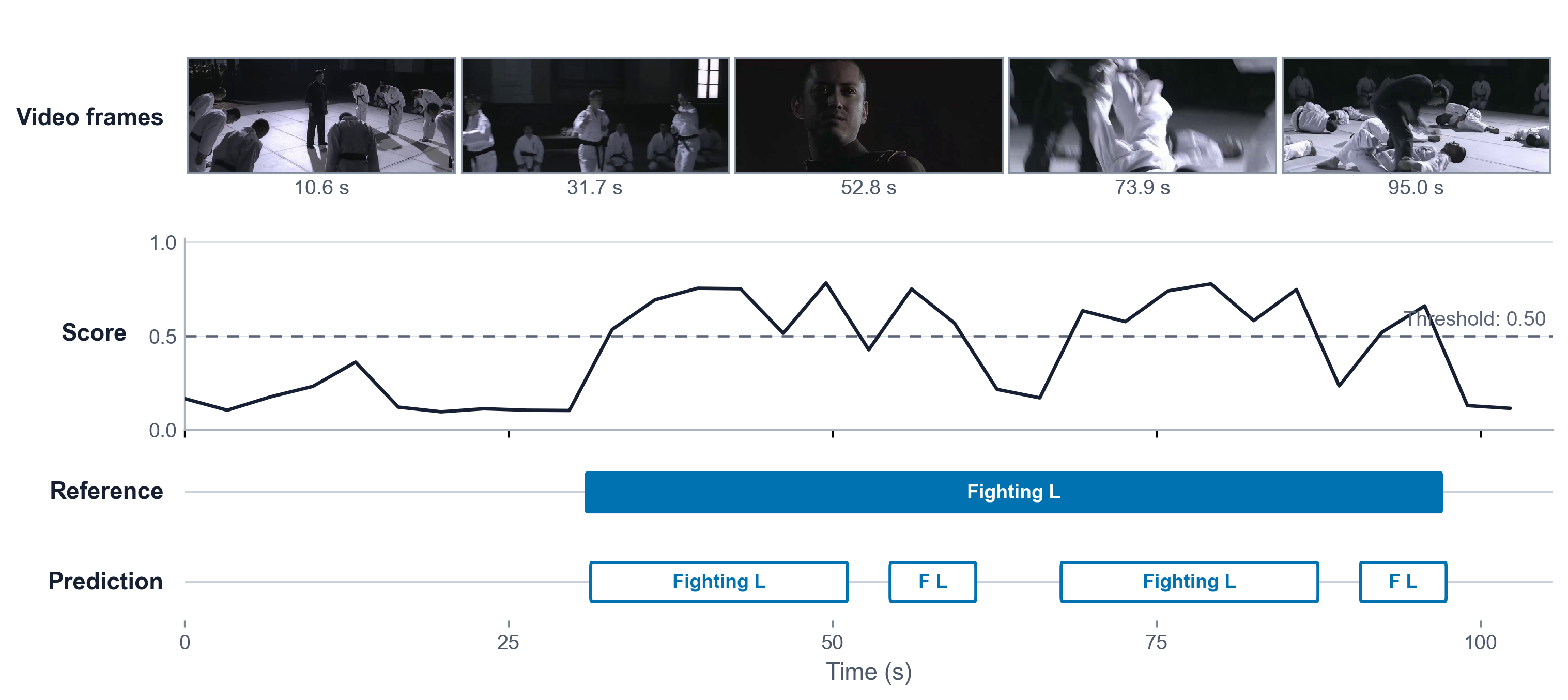}
    \caption{Qualitative temporal violence-localisation result for \textit{Ip Man} No.~1.}
    \label{fig:temporal_localisation_case_study_ip_man}
\end{figure*}

Figure~\ref{fig:temporal_localisation_case_study_ip_man} presents the temporal-localisation result for No.~1. The predicted violent intervals largely overlap with the reference fighting segment, indicating that the model successfully follows the extended fight over most of the scene. However, several short gaps remain within the predicted sequence. These gaps correspond to brief pauses between individual exchanges or moments when the camera moves away from the active fight, reducing the immediate visual evidence of violence. As a result, the localisation score temporarily falls below the decision threshold even though the broader scene remains part of the same violent event. This behaviour helps explain why the model can retrieve the correct general violent region while still producing fragmented predictions within long and continuously edited action sequences.

\subsubsection{\textit{John Wick}}
\label{sec:qual_john_wick}

Table~\ref{tab:john_wick_results} presents one of the more challenging cinematic cases considered in this study. The selected sequence contains dark illumination, flashing lights, rapid camera motion, and frequent changes in viewpoint, all of which can make individual visual observations difficult to interpret. Nevertheless, the multimodal model correctly distinguishes the main gun-related scenes at the video level. In No.~2--4, where actual shooting occurs, the visual, audio, and fused severity models all predict high violence, while the violence-type head correctly identifies \emph{shooting}. The agreement across modalities suggests that acoustic evidence, particularly gunshot-related cues, can provide stable complementary information when the visual stream is degraded by difficult cinematic presentation.

\begin{table*}[h]
\centering
\caption{John Wick: qualitative results.}
\label{tab:john_wick_results}

\begingroup
\footnotesize
\setlength{\tabcolsep}{3pt}
\renewcommand{\arraystretch}{1.12}

\begin{tabular}{
    @{}
    >{\centering\arraybackslash}m{0.60cm}
    >{\centering\arraybackslash}m{1.55cm}
    >{\centering\arraybackslash}m{0.80cm}
    >{\centering\arraybackslash}m{0.80cm}
    >{\centering\arraybackslash}m{1.15cm}
    >{\raggedright\arraybackslash}m{1.55cm}
    >{\raggedright\arraybackslash}m{4.20cm}
    >{\raggedright\arraybackslash}m{2.00cm}
    @{}
}
\toprule
\textbf{No.} &
\textbf{Timeline} &
\textbf{Visual} &
\textbf{Audio} &
\makecell{\textbf{Visual}\\\textbf{+ Audio}} &
\makecell[l]{\textbf{Violence}\\\textbf{types}} &
\makecell[l]{\textbf{Retrieved moments}\\
             \textbf{Relative interval (s) [peak score]}} &
\makecell[l]{\textbf{Human}\\\textbf{annotation}} \\
\midrule

1 & 00:59--01:09 & high & non & non & none &
\textemdash &
Gun threat \\
\cmidrule(lr){1-8}

2 & 04:17--04:24 & high & high & high & shooting &
\textemdash &
Shoot \\
\cmidrule(lr){1-8}

3 & 04:44--04:51 & high & high & high & shooting &
\textemdash &
Shoot \\
\cmidrule(lr){1-8}

4 & 04:53--05:08 & high & high & high & shooting &
\textemdash &
Shoot \\
\cmidrule(lr){1-8}

5 & 05:09--05:20 & non & high & high & shooting &
0.497--0.828 [0.5918] & Shoot, Dance \\

\bottomrule
\end{tabular}

\endgroup
\end{table*}

No.~1 follows a pattern similar to the gun-confrontation example observed in \textit{Hidden Man}. A character threatens another person with a gun, but no shot is fired. The visual-only model predicts high violence because the weapon itself provides a strong visual cue, whereas the audio-only model predicts non-violence in the absence of gunfire. After multimodal fusion, the scene is classified as non-violent, and the violence-type head predicts no violence type. This indicates that the acoustic stream can help distinguish the presence of a weapon from an actual shooting event and can suppress an overestimation produced by visual evidence alone.

However, a notable limitation is observed in temporal retrieval. Despite the correct video-level high-violence and \emph{shooting} predictions for No.~2--4, the localisation head retrieves no violent intervals. This contrast suggests that temporal localisation is more sensitive than video-level classification to rapid editing, dark lighting, flashing illumination, and frequent viewpoint changes. The global classification heads can integrate evidence over the complete clip, whereas the localisation head must determine whether individual sampled temporal regions contain sufficiently clear violent evidence. When the scene changes rapidly or the violent action is only briefly visible between cuts, the corresponding localisation score may fail to exceed the decision threshold. These results therefore suggest that accurate violence retrieval may require temporally clearer or more visually stable evidence than video-level violence recognition.

No.~5 further illustrates the difficulty of mixed-content scenes under rapid cinematic editing. The clip contains both shooting and dancing, and the audio and multimodal models classify it as high violence, while the violence-type head correctly identifies \emph{shooting}. The localisation head also retrieves a short violent interval near the beginning of the clip. This suggests that the model is able to capture the violent component even when it is embedded within a predominantly non-violent activity. At the same time, because the video-level severity and type heads summarise the complete clip with a single prediction, they do not explicitly represent the coexistence of violent and non-violent content within the same short segment. This case therefore highlights the value of temporal retrieval for identifying where the violent evidence occurs within mixed cinematic content.

\subsubsection{\textit{Mission: Impossible -- Fallout}}
\label{sec:qual_mission_impossible}

Table~\ref{tab:mission_impossible_fallout_results} contains several informative cases where the visual and audio streams provide conflicting evidence. In No.~3, the protagonist uses a laser scanner inside a toilet while people outside repeatedly knock on the door. The visual-only model predicts high violence, plausibly because the scanner has a gun-like appearance, whereas the audio-only model predicts low violence, with the repeated knocking providing an impact-like acoustic pattern that can resemble fighting. When the two modalities are combined, the severity prediction is correctly reduced to non-violence. In this clip, combining the streams corrects both unimodal severity predictions. However, the violence-type head still predicts \emph{shooting}, suggesting that the combination of the gun-like visual appearance and impact-like acoustic cues can produce an incorrect semantic interpretation even when the severity head correctly recognises the scene as non-violent.

\begin{table*}[h]
\centering
\caption{Mission: Impossible -- Fallout: qualitative results.}
\label{tab:mission_impossible_fallout_results}

\begingroup
\footnotesize
\setlength{\tabcolsep}{3pt}
\renewcommand{\arraystretch}{1.12}

\begin{tabular}{
    @{}
    >{\centering\arraybackslash}m{0.60cm}
    >{\centering\arraybackslash}m{1.55cm}
    >{\centering\arraybackslash}m{0.80cm}
    >{\centering\arraybackslash}m{0.80cm}
    >{\centering\arraybackslash}m{1.15cm}
    >{\raggedright\arraybackslash}m{1.55cm}
    >{\raggedright\arraybackslash}m{4.20cm}
    >{\raggedright\arraybackslash}m{2.00cm}
    @{}
}
\toprule
\textbf{No.} &
\textbf{Timeline} &
\textbf{Visual} &
\textbf{Audio} &
\makecell{\textbf{Visual}\\\textbf{+ Audio}} &
\makecell[l]{\textbf{Violence}\\\textbf{types}} &
\makecell[l]{\textbf{Retrieved moments}\\
             \textbf{Relative interval (s) [peak score]}} &
\makecell[l]{\textbf{Human}\\\textbf{annotation}} \\
\midrule

1 & 00:20--01:30 & low & low & low & fighting &
\makecell[l]{
33.809--38.172 [0.7140]\\
49.078--51.259 [0.5669]} &
Fight \\
\cmidrule(lr){1-8}

2 & 01:30--01:45 & non & non & non & none &
\textemdash &
Pee \\
\cmidrule(lr){1-8}

3 & 01:47--02:04 & high & low & non & shooting &
\textemdash &
Knock \\
\cmidrule(lr){1-8}

4 & 02:05--02:20 & high & low & non & fighting &
\textemdash &
Hug \\
\cmidrule(lr){1-8}

5 & 02:25--04:18 & low & low & low & fighting &
\makecell[l]{
72.198--79.242 [0.8129]\\
44.023--58.111 [0.7379]\\
1.761--33.458 [0.7062]} &
Fight, Club, Shoot \\
\cmidrule(lr){1-8}

6 & 04:42--05:05 & high & low & non & abuse &
\textemdash &
Talk \\
\cmidrule(lr){1-8}

7 & 05:07--05:23 & non & non & non & none &
\textemdash &
Talk \\

\bottomrule
\end{tabular}

\endgroup
\end{table*}

No.~4 shows a related pattern without the gun-like object. The scene contains no actual fight, but repeated physical and acoustic cues associated with the surrounding confrontation cause the visual-only and audio-only models to predict high and low violence, respectively. Multimodal fusion again reduces the severity prediction to non-violence, whereas the violence-type head predicts \emph{fighting}. Together, No.~3 and No.~4 indicate that the multimodal severity head can suppress ambiguous unimodal evidence, while the type head remains more sensitive to semantic patterns that resemble specific forms of violence.

No.~6 provides a different example of disagreement between the model outputs. In this scene, a man in the toilet has already lost consciousness and other characters search him and remove his belongings. Although no new physical attack occurs within the clip, the visual situation can resemble abuse when considered without the preceding narrative context. The violence-type head consequently predicts \emph{abuse}, while the multimodal severity head classifies the scene as non-violent. Similar to the \textit{Game of Thrones} example discussed earlier, this result shows that the type head can respond to the semantic appearance or aftermath of violence even when the severity head does not identify an active violent event.

\begin{figure*}[h]
    \centering
    \includegraphics[width=1\linewidth]{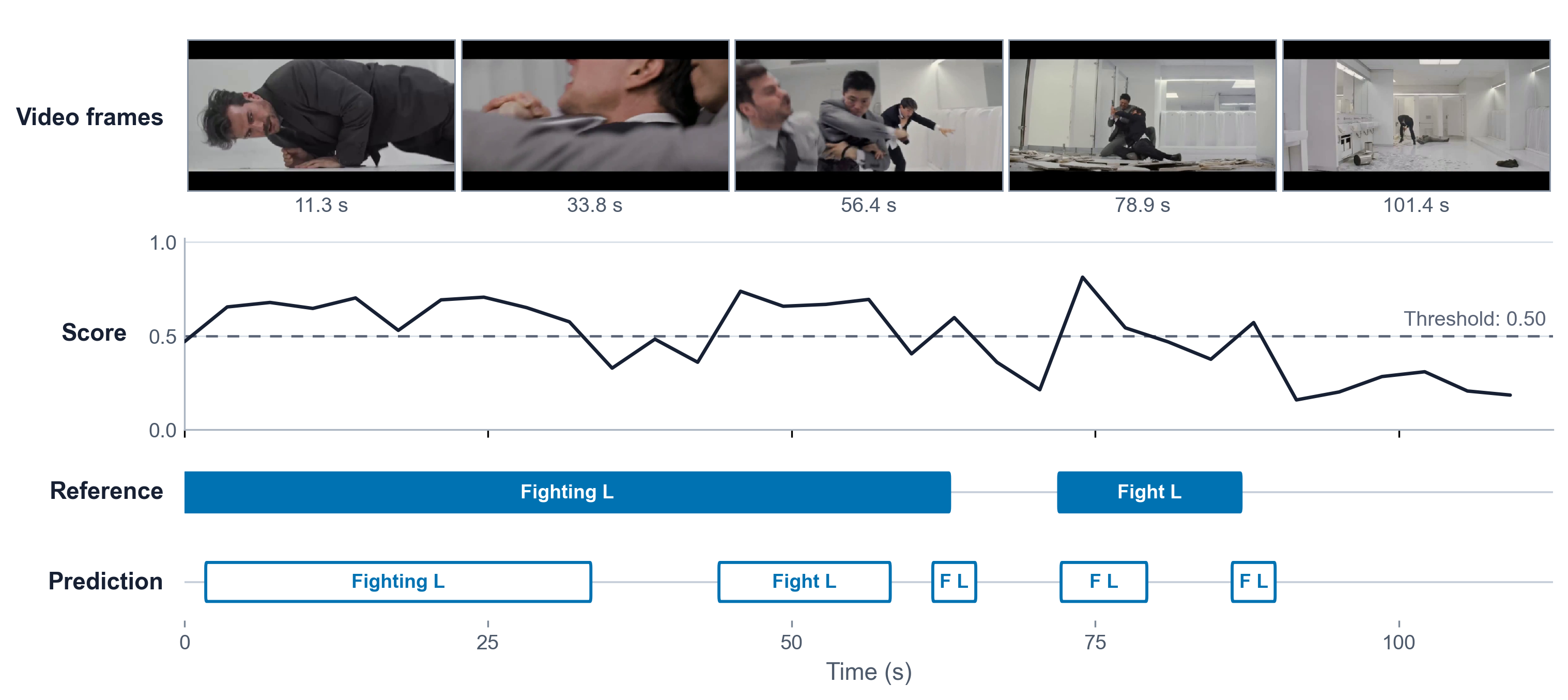}
    \caption{Qualitative temporal violence-localisation result for \textit{Mission: Impossible -- Fallout} No.~5.}
    \label{fig:temporal_localisation_case_study_mission_impossible}
\end{figure*}

Figure~\ref{fig:temporal_localisation_case_study_mission_impossible} further illustrates a limitation of temporal retrieval using No.~5. The model successfully identifies substantial portions of the extended fighting sequence, but the predicted intervals are fragmented, and some violent actions are missed. In particular, the brief gunshot is not retrieved. One likely reason is the limited temporal resolution of the current framework, where only $N=32$ temporal anchors are uniformly sampled across the complete video. For a relatively long clip, the temporal distance between adjacent anchors becomes larger, so a very brief action such as a gunshot may occur between sampled timestamps or provide insufficient evidence at the nearest sampled frame.

Another missed region occurs when one character continuously threatens another with a sharp object positioned around the neck. Although this action remains part of the broader violent encounter, it contains less dynamic motion than the surrounding fight. The localisation score can therefore fall below the decision threshold, separating what is semantically one continuous violent sequence into several predicted intervals. This result suggests that the current localisation head is more responsive to explicit and visually dynamic violent actions than to sustained threats or relatively static forms of violence. Together with the missed gunshot, the example highlights the dependence of temporal retrieval on both sampling resolution and the visual dynamics of the violent behaviour.

\subsubsection{\textit{The Boys}}
\label{sec:qual_the_boys}

Table~\ref{tab:the_boys_results} summarises the qualitative results for \textit{The Boys}. Because several scenes contain highly explicit violence, we report only the table-level analysis. No.~1--2 and No.~1--3 illustrate complementary modality behaviour. In No.~1--2, the visual model predicts low violence while the audio and fused models predict high violence; in No.~1--3, the visual model predicts high violence while the audio model predicts low violence, and fusion retains the high-violence decision. In both cases, the violence-type head identifies \emph{fighting} and \emph{shooting}. These examples show that multimodal fusion can preserve the stronger evidence from either modality depending on the scene.

\begin{table*}[h]
\centering
\caption{The Boys: qualitative results.}
\label{tab:the_boys_results}

\begingroup
\footnotesize
\setlength{\tabcolsep}{3pt}
\renewcommand{\arraystretch}{1.12}

\begin{tabular}{
    @{}
    >{\centering\arraybackslash}m{0.60cm}
    >{\centering\arraybackslash}m{1.55cm}
    >{\centering\arraybackslash}m{0.80cm}
    >{\centering\arraybackslash}m{0.80cm}
    >{\centering\arraybackslash}m{1.15cm}
    >{\raggedright\arraybackslash}m{1.55cm}
    >{\raggedright\arraybackslash}m{4.20cm}
    >{\raggedright\arraybackslash}m{2.00cm}
    @{}
}
\toprule
\textbf{No.} &
\textbf{Timeline} &
\textbf{Visual} &
\textbf{Audio} &
\makecell{\textbf{Visual}\\\textbf{+ Audio}} &
\makecell[l]{\textbf{Violence}\\\textbf{types}} &
\makecell[l]{\textbf{Retrieved moments}\\
             \textbf{Relative interval (s) [peak score]}} &
\makecell[l]{\textbf{Human}\\\textbf{annotation}} \\
\midrule

1-1 & 00:08--00:16 & low & non & non & none &
\textemdash &
Talk \\
\cmidrule(lr){1-8}

1-2 & 00:32--01:05 & low & high & high &
fighting, shooting &
\makecell[l]{
18.047--27.328 [0.8577]\\
29.391--33.000 [0.7280]} &
Fight, Shoot \\
\cmidrule(lr){1-8}

1-3 & 01:22--02:42 & high & low & high &
fighting, shooting &
\makecell[l]{
41.147--58.603 [0.8708]\\
36.159--38.653 [0.7659]\\
66.084--68.578 [0.6835]} &
Split, Fight, Shoot \\
\cmidrule(lr){1-8}

1-4 & 02:44--03:45 & low & low & low & fighting &
\makecell[l]{
8.616--39.249 [0.8411]\\
0.957--6.701 [0.7496]} &
Fight, Shoot \\
\cmidrule(lr){1-8}

1-5 & 03:52--05:06 & low & low & low & fighting &
\makecell[l]{
35.456--51.469 [0.8230]\\
5.719--28.594 [0.8039]\\
60.619--62.906 [0.5674]} &
Fight, Club \\
\cmidrule(lr){1-8}

1-6 & 05:08--05:15 & high & non & non & none &
\textemdash &
Talk \\
\cmidrule(lr){1-8}

1-7 & 05:25--05:32 & high & non & non & none &
\textemdash &
Talk \\
\cmidrule(lr){1-8}

1-8 & 05:34--05:51 & high & non & non & none &
\textemdash &
Run \\

\midrule

2-1 & 00:32--01:03 & non & non & non & none &
\textemdash &
Talk \\
\cmidrule(lr){1-8}

2-2 & 02:01--02:21 & non & non & non & none &
\textemdash &
Talk \\
\cmidrule(lr){1-8}

2-3 & 02:24--02:37 & non & non & non & none &
\textemdash &
Talk \\
\cmidrule(lr){1-8}

2-4 & 02:55--02:59 & non & non & non & none &
\textemdash &
Talk \\

\bottomrule
\end{tabular}

\endgroup
\end{table*}

Conversely, No.~1--6 to No.~1--8 are non-violent talking or running scenes for which the visual-only model predicts high violence, whereas the audio and fused models correctly predict non-violence. This suggests that acoustic information can suppress misleading visual cues inherited from surrounding explicit action scenes and improve discrimination between active violence and visually intense but non-violent content.

\subsubsection{\textit{The Punisher}}
\label{sec:qual_punisher}

Table~\ref{tab:the_punisher_results} shows that multimodal information is particularly useful when visually threatening cues are not accompanied by an actual violent event. For example, in No.~1 and No.~2, the visual-only model predicts high violence, whereas the audio and fused models reduce the prediction to non-violence. Similarly, in No.~9 the visual-only model predicts low violence, while multimodal fusion correctly reduces the prediction to non-violence. These cases again demonstrate how acoustic evidence can suppress misleading visual cues such as weapons, threatening posture, or the aftermath of nearby violent actions.

\begin{table*}[h]
\centering
\caption{The Punisher: qualitative results.}
\label{tab:the_punisher_results}

\begingroup
\footnotesize
\setlength{\tabcolsep}{3pt}
\renewcommand{\arraystretch}{1.12}

\begin{tabular}{
    @{}
    >{\centering\arraybackslash}m{0.60cm}
    >{\centering\arraybackslash}m{1.55cm}
    >{\centering\arraybackslash}m{0.80cm}
    >{\centering\arraybackslash}m{0.80cm}
    >{\centering\arraybackslash}m{1.15cm}
    >{\raggedright\arraybackslash}m{1.55cm}
    >{\raggedright\arraybackslash}m{4.20cm}
    >{\raggedright\arraybackslash}m{2.00cm}
    @{}
}
\toprule
\textbf{No.} &
\textbf{Timeline} &
\textbf{Visual} &
\textbf{Audio} &
\makecell{\textbf{Visual}\\\textbf{+ Audio}} &
\makecell[l]{\textbf{Violence}\\\textbf{types}} &
\makecell[l]{\textbf{Retrieved moments}\\
             \textbf{Relative interval (s) [peak score]}} &
\makecell[l]{\textbf{Human}\\\textbf{annotation}} \\
\midrule

1 & 00:03--00:12 & high & non & non & none &
\textemdash &
Gun threat \\
\cmidrule(lr){1-8}

2 & 00:20--00:27 & high & non & non & abuse &
\textemdash &
Call \\
\cmidrule(lr){1-8}

3 & 01:02--01:16 & high & high & high & shooting &
\makecell[l]{
4.804--7.310 [0.6717]\\
7.728--8.563 [0.5394]\\
3.551--3.968 [0.5053]} &
Shoot \\
\cmidrule(lr){1-8}

4 & 02:21--03:09 & high & low & high & shooting &
\textemdash &
Shoot \\
\cmidrule(lr){1-8}

5 & 03:34--04:28 & low & low & low & fighting &
26.043--27.723 [0.5014] &
Confront \\
\cmidrule(lr){1-8}

6 & 04:37--04:44 & low & low & low & fighting &
\textemdash &
Threat \\
\cmidrule(lr){1-8}

7 & 06:54--07:44 & low & low & low & fighting &
\makecell[l]{
0.783--27.398 [0.6867]\\
36.792--46.186 [0.6837]\\
47.752--50.100 [0.6136]} &
Fight, Stab \\
\cmidrule(lr){1-8}

8 & 09:16--09:30 & low & low & low & fighting &
0.206--1.856 [0.6184] &
Fight, Push \\
\cmidrule(lr){1-8}

9 & 09:59--10:05 & low & non & non & fighting &
\textemdash &
Threat \\

\bottomrule
\end{tabular}

\endgroup
\end{table*}

No.~5, No.~6, and No.~9 are also informative because their human annotations correspond to confrontation or threat rather than active violence, yet they occur immediately before or after violent scenes. The low-violence predictions for No.~5 and No.~6 are therefore understandable: the clips retain visual and contextual cues strongly associated with the surrounding fighting sequence even though no explicit violent action occurs within the selected interval. This illustrates the difficulty of assigning a discrete violence label to short cinematic clips when the boundary between threat, confrontation, and physical violence is gradual rather than clear.

\begin{figure*}[h]
    \centering
    \includegraphics[width=1\linewidth]{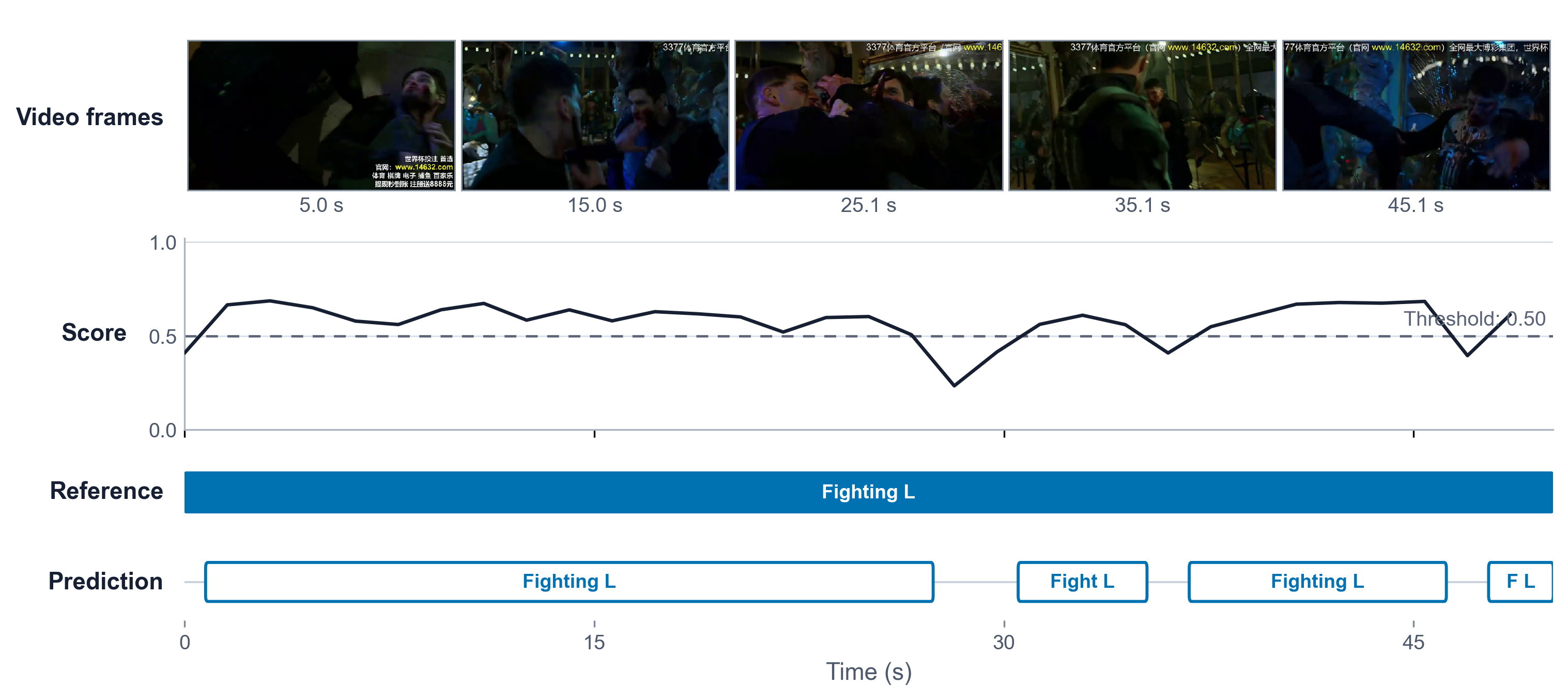}
    \caption{Qualitative temporal violence-localisation result for \textit{The Punisher} No.~7.}
    \label{fig:temporal_localisation_case_study_punisher}
\end{figure*}

Figure~\ref{fig:temporal_localisation_case_study_punisher} illustrates the temporal-localisation behaviour for No.~7. The reference annotation represents the sequence as one continuous violent event, whereas the model divides it into several predicted intervals. Most of the violent period is still retrieved, but short reductions in the localisation score cause the prediction to fall below the threshold and fragment the scene. As observed in the previous case studies, brief pauses, changes in camera viewpoint, or less dynamic moments within an otherwise continuous fight can weaken the immediate violence evidence and produce artificial temporal boundaries.

\subsubsection{Movie Violence and Official Ratings}
\label{sec:whole_movie_classification}

To broaden the comparison beyond the four feature films examined in the selected-scene case studies, we additionally include \textit{The Pursuit of Happyness} and \textit{Dunkirk}. The former provides a contrast with limited physical violence, while the latter introduces wartime shooting, explosions, and sustained threat. These contrasting portrayals help examine how the movie-level profiles relate to official classification advice across different forms and amounts of violent content. We extended the selected-scene case studies to complete inference on six feature films. We partitioned each movie using the pretrained image-based variant of the local-to-global scene segmentation framework (LGSS) \citep{rao2020local}. We subdivided the long scenes into clips of at most 119~s, and ran inference for every analysis clip, providing complete temporal coverage of the six movies.


We summarise each movie using a compact profile that captures the proportion of runtime classified as violent, exposure to violence in high-labelled clips, and the most frequently predicted violence types. We map clip-relative intervals onto the original movie timeline and merge overlapping intervals. We calculate the retrieved violence proportion by dividing the total retrieved violent seconds by the movie’s full runtime. We calculate high-violence exposure by dividing the retrieved violent seconds within high-labelled clips by the full runtime. We rank violence types by the number of distinct scenes in which the model predicts each type. Table~\ref{tab:whole_movie_classification} reports both exposure measures and the most frequent types alongside official age classifications and violence-related advice.

\begin{table*}[h]
\centering
\caption{Whole-movie violence summary and official classification advice. Official advice is paraphrased from the BBFC records or translated from the BAMID classification justification.}
\label{tab:whole_movie_classification}

\begingroup
\small
\setlength{\tabcolsep}{3pt}
\renewcommand{\arraystretch}{1.18}

\begin{tabularx}{\textwidth}{
    @{}
    >{\raggedright\arraybackslash}p{0.18\textwidth}
    >{\centering\arraybackslash}p{0.09\textwidth}
    >{\centering\arraybackslash}p{0.09\textwidth}
    >{\raggedright\arraybackslash}p{0.14\textwidth}
    >{\centering\arraybackslash}p{0.14\textwidth}
    >{\raggedright\arraybackslash}X
    @{}
}
\toprule
\textbf{Film} &
\textbf{Violence (\%)} &
\textbf{High-violence (\%)} &
\textbf{Most frequent predicted types} &
\textbf{Official classification} &
\textbf{Advice from official classification} \\
\midrule

\textit{The Pursuit of Happyness} (2006) &
0.29 &
0.00 &
Fighting &
BBFC 12 \citeyearpar{bbfc_happyness} &
Mild scuffling; strong language \\

\textit{Hidden Man} (2018) &
5.49 &
1.17 &
Fighting, shooting &
BAMID 15+ \citeyearpar{bamid_hiddenman} &
Crime, violence and killing \\

\textit{Dunkirk} (2017) &
5.86 &
3.97 &
Explosion, shooting &
BBFC 12 \citeyearpar{bbfc_dunkirk} &
Moderate violence; sustained wartime threat \\

\textit{Mission: Impossible - Fallout} (2018) &
6.89 &
4.42 &
Fighting, shooting &
BBFC 12 \citeyearpar{bbfc_fallout} &
Moderate violence; injury detail \\

\textit{John Wick} (2014) &
8.44 &
4.57 &
Fighting, shooting &
BBFC 15 \citeyearpar{bbfc_johnwick} &
Strong violence; bloody detail; beatings, stabbings, and gunplay \\

\textit{Ip Man} (2008) &
10.63 &
0.00 &
Fighting, abuse &
BBFC 15 \citeyearpar{bbfc_ipman} &
Moderate martial arts violence; bloody injury \\

\bottomrule
\end{tabularx}

\par\smallskip
\begin{minipage}{\textwidth}
\footnotesize
\textit{Notes.} BBFC denotes the British Board of Film Classification (UK); BAMID denotes Taiwan's Bureau of Audiovisual and Music Industry Development. A value of 0.00 indicates no retrieved violence within high-labelled clips.
\end{minipage}

\endgroup
\end{table*}

The profiles show broad qualitative correspondence with the official content descriptions. \textit{The Pursuit of Happyness} has the smallest retrieved violence proportion, consistent with the BBFC's description of a mild scuffle. Fighting is the most frequent prediction in \textit{Ip Man}, corresponding to its martial arts content, while the frequent fighting and shooting predictions in \textit{John Wick} and \textit{Fallout} reflect forms of violence described in their official advice. Similarly, the explosion and shooting predictions for \textit{Dunkirk} correspond to its wartime action. For \textit{Hidden Man}, the fighting and shooting predictions are broadly consistent with the Taiwanese justification, which explicitly identifies crime, violence and killing as grounds for its classification.

However, retrieved duration alone does not determine an age rating. Although \textit{Ip Man} has a larger retrieved proportion than \textit{John Wick}, their BBFC advice describes moderate martial arts violence and strong violence, respectively. Moreover, the BBFC 12 record for \textit{The Pursuit of Happyness} highlights strong language despite its much smaller retrieved violence proportion. These differences reflect the role of presentation, frequency, detail and narrative context in classification \citep{bbfc_violence_guide}. The proposed summaries can therefore provide interpretable evidence for human review by describing the temporal extent, operational category composition, and forms of predicted violence; they do not establish a calibrated mapping from model outputs to official age categories.

\section{Discussion}
\subsection{Summary of Results}
We presented a multimodal framework for understanding and retrieving violence in films and streaming media.     Our  CLIP+BEATs model achieved 93.27\% accuracy and a
macro-F1 of 0.9208 for violence-severity classification on
XD-Violence. For violence-type prediction, it achieved a
micro-F1 of 0.8860 and the highest F1 for five of the six
categories. These results support the use of complementary audio
and visual evidence for violence detection. 

The contribution of each modality also varies by task. BEATs
achieves stronger standalone performance than CLIP for severity
classification, whereas CLIP performs better for violence-type
prediction. This difference suggests that acoustic evidence can
help characterise the intensity of an event, while visual
information provides useful cues about its semantic category.
However, abuse remains difficult to classify, indicating that
some categories depend on contextual information that short
audio-visual segments may not adequately capture.

The cross-domain experiment achieves an accuracy of 0.6057 and
a macro-F1 of 0.5834 on AIRTLab, compared with 0.7886 accuracy
and 0.7921 macro-F1 after in-domain training. This difference
indicates that performance depends on the relationship between
the training and evaluation domains. The results motivate
further testing across different recording conditions, scene
types, and production settings before the framework can be
considered robust across domains.

The qualitative analysis of representative movie and television scenes provides further evidence of the complementary role of the two modalities. Acoustic information frequently corrected misleading visual cues, such as the presence of weapons or visually threatening situations without an accompanying violent action, while visual information remained important for identifying the semantic form of violence. The case studies also show that different prediction heads capture different aspects of a scene: severity, violence type, and temporal localisation do not always agree when violence is ambiguous, context-dependent, or mixed with non-violent activity. Temporal retrieval was generally able to identify the main violent portions of longer scenes, but short actions and continuous violent sequences with pauses or rapid cinematic changes remained challenging.

The whole-movie profiles provide a complementary view of
predicted violence by summarising retrieved duration,
high-violence exposure, and the most frequent predicted
violence types. \cite{martinez2019violence}
studied violence-rating prediction from movie scripts, using
textual information rather than directly retrieving violent
moments from audio-visual content. Our approach complements
script-based analysis by identifying predicted violent intervals
and their associated types in the film itself.

The movie-level results show broad qualitative correspondence
between predicted violence profiles and official classification
advice. However, this correspondence should not be interpreted
as evidence that the model can determine age ratings. The
British Board of Film Classification considers factors such as
context, frequency, detail, and presentation when assessing
violence \citep{bbfc2025violence}. Retrieved duration alone cannot
represent all these considerations. The proposed profiles are
therefore best used to support human reviewers by identifying
potentially relevant scenes, rather than replacing expert
judgement or official classification procedures.

In relation to the literature, \cite{wu2020not} investigated multimodal violence
detection under weak supervision and demonstrated the value of
combining visual and acoustic information. Our results extend
this line of work by integrating CLIP and BEATs representations
within a shared framework that predicts violence severity,
violence type, and temporal location. The improvements over the
unimodal configurations indicate that these representations
provide complementary evidence in our experimental setting.

\subsection{Limitations and Future Directions}
\label{sec:limitations}

\subsubsection{Context-Dependent Violence-Type Recognition}

Context-dependent categories are difficult to distinguish from threats, restraint, injury, and the aftermath of violence. We found that subtle and context-dependent violence types remain difficult to recognise reliably. As shown in Section~\ref{sec:type_results}, abuse achieves substantially lower and more variable performance than the other violence categories, and multimodal fusion does not improve this category over the visual-only model. The qualitative results further show that the violence-type head can respond to the semantic appearance of violence even when no active violent event is occurring. For example, in \textit{Game of Thrones} No.~2--2, the severity and localisation outputs indicate non-violence, whereas the type head predicts \emph{abuse} because the isolated scene contains an injured, blood-covered character being physically held. Similar ambiguity appears in \textit{Mission: Impossible -- Fallout} No.~6, where the aftermath of an earlier confrontation is interpreted as abuse. These cases suggest that short video excerpts may not provide sufficient narrative context to distinguish an abusive action from its aftermath, restraint, injury, or other visually similar situations. 

Future work could incorporate longer-range narrative context and additional textual information, such as subtitles or dialogue, to provide semantic evidence that is not directly available from visual and acoustic cues. Such an extension may be particularly useful for context-dependent categories such as abuse, where the interpretation of an isolated scene often depends on preceding or surrounding events.

\subsubsection{Sensitivity to Cinematic Context}

Cinematic context can produce plausible but incorrect predictions. The qualitative analysis in Section~\ref{sec:qualitative_results} shows that the model may confuse confrontation, threat, anticipation, or visually similar motion with active violence. Examples include sword-drawing being interpreted as fighting in \textit{Game of Thrones}, gun confrontation being treated as high violence by the visual model in \textit{Hidden Man}, running and rolling being confused with a car accident, and a laser scanner combined with repeated door-knocking cues being interpreted as shooting in \textit{Mission: Impossible -- Fallout}. Multimodal fusion often reduces these errors by combining complementary visual and acoustic evidence, but it does not eliminate them. This indicates that the current framework remains sensitive to cinematic cues that are strongly associated with violence but do not necessarily correspond to an actual violent act.

Future work could therefore investigate stronger cross-modal reasoning and training with challenging cinematic examples that explicitly distinguish active violence from threats, preparation, aftermath, and visually similar non-violent actions. This may help the model learn not only whether violence-related cues are present, but also whether those cues correspond to an actual violent event.

\subsubsection{Fragmentation of Continuous Violent Events}

Continuous violent events can be fragmented into multiple retrieved intervals. Although the temporal-localisation results in Section~\ref{sec:localisation_results} show that the model can identify the approximate regions containing violence, the qualitative examples reveal that long fighting sequences are sometimes split when the localisation score briefly falls below the decision threshold. This behaviour is visible in \textit{Ip Man}, \textit{Mission: Impossible -- Fallout}, and \textit{The Punisher}, where pauses between attacks, camera changes, or less dynamic threatening actions create short gaps within otherwise continuous violent scenes. Relatively static actions, such as holding a sharp object near a character's neck, may also receive lower localisation scores than visually dynamic actions such as punching or fighting. 

Future work could explore higher-resolution or adaptive temporal sampling together with continuity-aware temporal modelling or post-processing. Rather than treating every short drop below the localisation threshold as a genuine event boundary, such methods could use neighbouring temporal evidence to preserve semantically continuous violent events while still separating genuinely distinct incidents.


\section{Conclusions}
\label{sec:conclusions}


We developed a multimodal framework for movie-violence understanding that combines CLIP visual features and BEATs audio features through adaptive fusion and Transformer-based temporal modelling. Our framework jointly supports violence-severity classification, multi-label violence-type prediction, and temporal retrieval, providing a unified approach to analysing violent content in long-form movies.

On XD-Violence, our fused model achieved 93.27\% accuracy for severity classification and a micro-F1 score of 0.8860 for violence-type prediction. We achieved a mean temporal intersection-over-union (tIoU) of 0.5075 for temporal localisation, demonstrating that the framework can identify approximate violent regions, although precise temporal boundaries remain challenging. Our results also show that the two modalities contribute differently across tasks, supporting the value of integrating audio and visual evidence rather than relying on either source in isolation. Our CLIP-based cross-domain evaluation on AIRTLab revealed a substantial domain gap, highlighting the need for broader testing before we can establish the framework's reliability across different datasets and cinematic styles.


Beyond classification performance, our framework can help reviewers locate relevant scenes without manually screening every minute of a film. These profiles may also support comparative analysis of violence exposure across movies and provide a starting point for more detailed investigation. However, predicted violence duration and category labels do not capture every factor involved in content assessment, including narrative context, the nature of the threat, the consequences of an act, and how violence is presented to the audience. Therefore, we position the framework as a decision-support tool for human reviewers rather than an automated film-rating system.

Our findings demonstrate that effective movie-violence analysis requires more than detecting visually dramatic actions. It requires combining complementary audio-visual cues with temporal and contextual reasoning. We demonstrate the potential of multimodal learning for this task while showing that detecting an event, identifying its type, and determining its precise temporal extent remain distinct challenges. In particular, context-dependent categories, brief violent actions, and changes in the evaluation domain continue to limit performance.



Future work can investigate narrative context through dialogue and subtitles, adaptive temporal sampling, and continuity-aware localisation. We can also evaluate the framework on additional independent datasets and use systematically sampled scenes with annotations from multiple reviewers to establish its robustness and practical utility. These extensions may help us improve the identification of active violence, distinguish it from threats, preparation and aftermath, and retrieve continuous violent events more accurately.
\section*{Code Availability}

The source code and supporting implementation materials for this study are publicly available in the Sydney Machine Learning GitHub repository: \url{https://github.com/sydney-machine-learning/multimodal-movieanalysis}.


\appendix

\section{Implementation Details}
\label{app:implementation_details}

All models use $N=32$ uniformly sampled temporal positions, hidden dimension $H=256$, 2 Transformer encoder layers with 8 attention heads, a feed-forward dimension of 1024, and dropout of 0.1. The temporal-attention pooling module uses dimension 128. Pretrained CLIP and BEATs encoders are frozen; training uses cached features.

\begin{table}[h]
\centering
\caption{Training and preprocessing settings.}
\label{tab:implementation_details}
\footnotesize
\setlength{\tabcolsep}{3pt}
\renewcommand{\arraystretch}{1.08}

\begin{tabularx}{\columnwidth}{
    @{}
    >{\raggedright\arraybackslash}p{0.23\columnwidth}
    >{\raggedright\arraybackslash}X
    @{}
}
\toprule
\textbf{Setting} & \textbf{Configuration} \\
\midrule

Optimisation &
AdamW; weight decay 0.01. LR: $3\times10^{-4}$ (video level);
$10^{-5}$ (shared layers), $10^{-4}$ (localisation head). \\

Batch size &
32 (CLIP); 64 (BEATs, CLIP+BEATs). \\

Training &
Video level: max 50 epochs, patience 7 on validation macro-F1.
Localisation: max 30 epochs, selected by validation mean tIoU. \\

Class imbalance &
No weighting for severity; inverse-frequency sampling and positive BCE
weights for retrieval, capped at 20. \\

Loss weights &
$\lambda_{\mathrm{type}}=0.5$,
$\lambda_{\mathrm{loc}}=1.0$,
$\lambda_{\mathrm{sev}}=0.1$,
$\lambda_{\mathrm{cons}}=0.1$. \\

Thresholds &
Type threshold selected by validation macro-F1;
$\tau_{\mathrm{loc}}=0.5$. \\

Preprocessing &
RGB: $336\times336$, CLIP processor, $\ell_2$ normalisation.
Audio: $\pm1$ s, 16 kHz, frozen BEATs. \\

\bottomrule
\end{tabularx}
\end{table}



\bibliographystyle{elsarticle-harv} 
\bibliography{cleaned_references_download}

@article{wu2020not,
  author        = {Wu, Peng and Liu, Jing and Shi, Yujia and Sun, Yujia and
                   Shao, Fangtao and Wu, Zhaoyang and Yang, Zhiwei},
  title         = {Not Only Look, but Also Listen: Learning Multimodal
                   Violence Detection under Weak Supervision},
  journal       = {arXiv preprint arXiv:2007.04687},
  year          = {2020},
  eprint        = {2007.04687},
  archivePrefix = {arXiv},
  primaryClass  = {cs.CV},
  url           = {https://arxiv.org/abs/2007.04687}
}

@article{wu2023weakly,
  author  = {Wu, Peng and Liu, Xiaotao and Liu, Jing},
  title   = {Weakly Supervised Audio-Visual Violence Detection},
  journal = {IEEE Transactions on Multimedia},
  volume  = {25},
  pages   = {1674--1685},
  year    = {2023},
  doi     = {10.1109/TMM.2022.3147369},
  url     = {https://doi.org/10.1109/TMM.2022.3147369}
}

@inproceedings{martinez2019violence,
  author    = {Martinez, Victor R. and Somandepalli, Krishna and
               Singla, Karan and Ramakrishna, Anil and Uhls, Yalda T. and
               Narayanan, Shrikanth},
  title     = {Violence Rating Prediction from Movie Scripts},
  booktitle = {Proceedings of the AAAI Conference on Artificial Intelligence},
  volume    = {33},
  number    = {1},
  pages     = {671--678},
  year      = {2019},
  doi       = {10.1609/aaai.v33i01.3301671},
  url       = {https://doi.org/10.1609/aaai.v33i01.3301671}
}

@misc{bbfc2025violence,
  author       = {{British Board of Film Classification}},
  title        = {{BBFC Guide -- Violence}},
  year         = {2025},
  howpublished = {\url{https://www.bbfc.co.uk/parents-guide-age-ratings/bbfc-guide-violence}},
  note         = {Published 25 March 2025. Accessed 8 October 2026}
}

@article{sands2026evaluation,
  author  = {Sands, Brendan and Wang, Yining and Xu, Chenhao and
             Zhou, Yuxuan and Wei, Lai and Chandra, Rohitash},
  title   = {An Evaluation of LLMs for Generating Movie Reviews:
             GPT-4o, Gemini-2.0 and DeepSeek-V3},
  journal = {Neural Computing and Applications},
  volume  = {38},
  number  = {13},
  pages   = {556},
  year    = {2026},
  doi     = {10.1007/s00521-026-12247-0},
  url     = {https://doi.org/10.1007/s00521-026-12247-0}
}

@article{perperis2011multimodal,
  author  = {Perperis, Thanassis and Giannakopoulos, Theodoros and
             Makris, Alexandros and Kosmopoulos, Dimitrios I. and
             Tsekeridou, Sofia and Perantonis, Stavros J. and
             Theodoridis, Sergios},
  title   = {Multimodal and Ontology-Based Fusion Approaches of Audio
             and Visual Processing for Violence Detection in Movies},
  journal = {Expert Systems with Applications},
  volume  = {38},
  number  = {11},
  pages   = {14102--14116},
  year    = {2011},
  doi     = {10.1016/j.eswa.2011.04.219},
  url     = {https://doi.org/10.1016/j.eswa.2011.04.219}
}

@misc{tirupattur2021violence,author        = {Tirupattur, Praveen and Schulze, Christian
                   and Dengel, Andreas},
  title         = {Violence Detection in Videos},
  year          = {2021},
  eprint        = {2109.08941},
  archivePrefix = {arXiv},
  primaryClass  = {cs.CV},
  url           = {https://arxiv.org/abs/2109.08941}
}

@article{peixoto2021harnessing,author  = {Peixoto, Bruno M. and Lavi, Bahram and Dias, Zanoni
             and Rocha, Anderson},
  title   = {Harnessing High-Level Concepts, Visual, and
             Auditory Features for Violence Detection in Videos},
  journal = {Journal of Visual Communication and Image Representation},
  volume  = {78},
  pages   = {103174},
  year    = {2021},
  doi     = {10.1016/j.jvcir.2021.103174}
}

@inproceedings{clarin2005dove,title={DOVE: Detection of movie violence using computer vision},
  author={Clarin, T. and Dionisio, J. and Echavez, M. and Naval, P. C.},
  booktitle={European Conference on Computer Vision Workshops},
  pages={150--156},
  year={2005}
}

@article{chen2011violence,title={Violent scene detection in movies},
  author={Chen, L.-H. and Su, C.-W. and Hsu, H.-W.},
  journal={International Journal of Pattern Recognition and Artificial Intelligence},
  volume={25},
  number={8},
  pages={1161--1172},
  year={2011},
  doi={10.1142/S0218001411009056}
}

@article{chandra2025music,title={Language models for longitudinal analysis of abusive content in Billboard Music Charts},
  author={Chandra, Rohitash and Suresh, Yathin and Sinha, Divyansh Raj and Jindal, Sanchit},
  journal={arXiv preprint arXiv:2510.06266},
  year={2025}
}

@article{bushman2013gun,title={Gun violence trends in movies},
author={Bushman, Brad J. and Jamieson, Patrick E. and Weitz, Ilana and Romer, Daniel},
journal={Pediatrics},
volume={132},
number={6},
pages={1014--1018},
year={2013},
doi={10.1542/peds.2013-1600}
}

@article{romer2017continuing,title={The continuing rise of gun violence in {PG-13} movies, 1985 to 2015},
author={Romer, Daniel and Jamieson, Patrick E. and Jamieson, Kathleen Hall},
journal={Pediatrics},
volume={139},
number={2},
pages={e20162891},
year={2017},
doi={10.1542/peds.2016-2891}
}

@article{fotouhi2025trends,
  author  = {Fotouhi, Babak and Tohidi, Amir and Touserkani, Rouzbeh and
             Bushman, Brad J.},
  title   = {Trends of Violence in Movies During the Past Half Century},
  journal = {JAMA Pediatrics},
  volume  = {179},
  number  = {3},
  pages   = {346--348},
  year    = {2025},
  doi     = {10.1001/jamapediatrics.2024.5741},
  url     = {https://doi.org/10.1001/jamapediatrics.2024.5741}
}

@article{huesmann2007impact,title={The impact of electronic media violence: scientific theory and research},
author={Huesmann, L. Rowell},
journal={Journal of Adolescent Health},
volume={41},
number={6},
pages={S6--S13},
year={2007},
doi={10.1016/j.jadohealth.2007.09.005}
}

@article{worth2008exposure,title={Exposure of {US} adolescents to extremely violent movies},
author={Worth, Keilah A. and Chambers, John G. and Nassau, Deborah H. and Rakhra, Balvinder K. and Sargent, James D.},
journal={Pediatrics},
volume={122},
number={2},
pages={306--312},
year={2008}
}

@article{coker2015media,title={Media violence exposure and physical aggression in fifth-grade children},
author={Coker, Tumaini R. and Elliott, Marc N. and Schwebel, David C. and Windle, Michael and Toomey, Sara L. and Tortolero, Susan R. and Hertz, Marci F. and Peskin, Melissa F. and Schuster, Mark A.},
journal={Academic Pediatrics},
volume={15},
number={1},
pages={82--88},
year={2015},
doi={10.1016/j.acap.2014.09.008}
}

@article{dillon2017effects,title={Effects of exposure to gun violence in movies on children's interest in real guns},
author={Dillon, Kelly P. and Bushman, Brad J.},
journal={JAMA Pediatrics},
volume={171},
number={11},
pages={1057--1062},
year={2017},
doi={10.1001/jamapediatrics.2017.2229}
}

@article{romer2014parental,title={Parental desensitization to violence and sex in movies},
author={Romer, Daniel and Jamieson, Patrick E. and Bushman, Brad J. and Bleakley, Amy and Wang, Anli and Langleben, Daniel and Jamieson, Kathleen Hall},
journal={Pediatrics},
volume={134},
number={5},
pages={877--884},
year={2014},
doi={10.1542/peds.2014-1167}
}

@article{romer2018parental,title={Parental desensitization to gun violence in {PG-13} movies},
author={Romer, Daniel and Jamieson, Patrick E. and Jamieson, Kathleen Hall},
journal={Pediatrics},
volume={141},
number={6},
pages={e20173491},
year={2018},
doi={10.1542/peds.2017-3491}
}

@misc{mpa_ratings,title={{Film Ratings}},
author={{Motion Picture Association}},
year={2026},
note={Available at: \url{https://www.motionpictures.org/film-ratings/}}
}

@misc{bbfc_guidelines,title={{BBFC Age Ratings and Classification Guidelines}},
author={{British Board of Film Classification}},
year={2026},
note={Available at: \url{https://www.bbfc.co.uk/}}
}

@misc{australian_classification,title={{What are the ratings?}},
author={{Australian Classification}},
year={2026},
note={Available at: \url{https://www.classification.gov.au/classification-ratings/what-are-ratings}}
}

@misc{guardian2026bbfc,title={The Pitt and Game of Thrones spinoff given age ratings as {BBFC} deploys new {AI} tool},
author={{The Guardian}},
year={2026},
note={Available at: \url{https://www.theguardian.com/tv-and-radio/2026/apr/15/game-of-thrones-euphoria-hbo-max-uk-age-ratings-bbfc-ai-tool}}
}

@inproceedings{lin2009violence,
  author    = {Lin, Jian and Wang, Weiqiang},
  title     = {Weakly-Supervised Violence Detection in Movies with
               Audio and Video Based Co-Training},
  booktitle = {Advances in Multimedia Information Processing -- PCM 2009},
  pages     = {930--935},
  year      = {2009},
  publisher = {Springer},
  doi       = {10.1007/978-3-642-10467-1_84},
  url       = {https://doi.org/10.1007/978-3-642-10467-1_84}
}

@article{eyben2013affective,
  author  = {Eyben, Florian and Weninger, Felix and Lehment, Nicolas and
             Schuller, Bjorn and Rigoll, Gerhard},
  title   = {Affective Video Retrieval: Violence Detection in Hollywood
             Movies by Large-Scale Segmental Feature Extraction},
  journal = {PLOS ONE},
  volume  = {8},
  number  = {12},
  pages   = {e78506},
  year    = {2013},
  doi     = {10.1371/journal.pone.0078506},
  url     = {https://doi.org/10.1371/journal.pone.0078506}
}

@inproceedings{pang2021violence,
  author    = {Pang, Wen-Feng and He, Qian-Hua and Hu, Yongjian and
               Li, Yan-Xiong},
  title     = {Violence Detection in Videos Based on Fusing Visual and
               Audio Information},
  booktitle = {Proceedings of the IEEE International Conference on
               Acoustics, Speech and Signal Processing (ICASSP)},
  pages     = {2260--2264},
  year      = {2021},
  publisher = {IEEE},
  doi       = {10.1109/ICASSP39728.2021.9413686},
  url       = {https://doi.org/10.1109/ICASSP39728.2021.9413686}
}

@inproceedings{sharma2022cnn,
  author    = {Sharma, Mayank and Gupta, Tarun and Qiu, Kenny and
               Hao, Xiang and Hamid, Raffay},
  title     = {CNN-Based Audio Event Recognition for Automated Violence
               Classification and Rating for Prime Video Content},
  booktitle = {Proceedings of Interspeech 2022},
  pages     = {2758--2762},
  year      = {2022},
  doi       = {10.21437/Interspeech.2022-10053},
  url       = {https://doi.org/10.21437/Interspeech.2022-10053}
}

@article{martinez2020victim,
  author        = {Martinez, Victor R. and Somandepalli, Krishna and
                   Singla, Karan and Ramakrishna, Anil and Uhls, Yalda T. and
                   Narayanan, Shrikanth},
  title         = {Victim or Perpetrator? Analysis of Violent Character
                   Portrayals from Movie Scripts},
  journal       = {arXiv preprint arXiv:2008.08225},
  year          = {2020},
  eprint        = {2008.08225},
  archivePrefix = {arXiv},
  primaryClass  = {cs.CL},
  url           = {https://arxiv.org/abs/2008.08225}
}

@article{na2024leveraging,
  author  = {Na, Gwangho and Ko, Jaepil and Cheoi, Kyungjoo},
  title   = {Leveraging Multi-Modality and Enhanced Temporal Networks
             for Robust Violence Detection},
  journal = {Machine Learning and Knowledge Extraction},
  volume  = {6},
  number  = {4},
  pages   = {2422--2434},
  year    = {2024},
  doi     = {10.3390/make6040119},
  url     = {https://doi.org/10.3390/make6040119}
}

@inproceedings{lin2010violence,author    = {Lin, Jian and Sun, Yi and Wang, Weiqiang},
  title     = {Violence Detection in Movies with Auditory and Visual Cues},
  booktitle = {Proceedings of the International Conference on Computational
               Intelligence and Security},
  pages     = {561--565},
  year      = {2010},
  publisher = {IEEE},
  doi       = {10.1109/CIS.2010.128}
}

@inproceedings{hassner2012violent,author    = {Hassner, Tal and Itcher, Yossi and Kliper-Gross, Orit},
  title     = {Violent Flows: Real-Time Detection of Violent Crowd Behavior},
  booktitle = {Proceedings of the IEEE Computer Society Conference on Computer
               Vision and Pattern Recognition Workshops},
  pages     = {1--6},
  year      = {2012},
  publisher = {IEEE},
  doi       = {10.1109/CVPRW.2012.6239348}
}

@article{gao2016violence,author  = {Gao, Yuan and Liu, Hong and Sun, Xiaohu and Wang, Can and Liu, Yi},
  title   = {Violence Detection Using Oriented Violent Flows},
  journal = {Image and Vision Computing},
  volume  = {48--49},
  pages   = {37--41},
  year    = {2016},
  doi     = {10.1016/j.imavis.2016.01.006}
}

@article{zhou2018violence,author  = {Zhou, Peipei and Ding, Qinghai and Luo, Haibo and Hou, Xinglin},
  title   = {Violence Detection in Surveillance Video Using Low-Level Features},
  journal = {PLOS ONE},
  volume  = {13},
  number  = {10},
  pages   = {e0203668},
  year    = {2018},
  doi     = {10.1371/journal.pone.0203668}
}

@inproceedings{tran2015learning,author    = {Tran, Du and Bourdev, Lubomir and Fergus, Rob and
               Torresani, Lorenzo and Paluri, Manohar},
  title     = {Learning Spatiotemporal Features with 3D Convolutional Networks},
  booktitle = {Proceedings of the IEEE International Conference on Computer Vision},
  pages     = {4489--4497},
  year      = {2015},
  publisher = {IEEE},
  doi       = {10.1109/ICCV.2015.510}
}

@inproceedings{sudhakaran2017learning,author    = {Sudhakaran, Swathikiran and Lanz, Oswald},
  title     = {Learning to Detect Violent Videos Using Convolutional
               Long Short-Term Memory},
  booktitle = {Proceedings of the 14th IEEE International Conference on
               Advanced Video and Signal Based Surveillance},
  pages     = {1--6},
  year      = {2017},
  publisher = {IEEE},
  doi       = {10.1109/AVSS.2017.8078468}
}

@article{ullah2019violence,author  = {Ullah, Fath U. Min and Ullah, Amin and Muhammad, Khan and
             Haq, Ijaz Ul and Baik, Sung Wook},
  title   = {Violence Detection Using Spatiotemporal Features with
             3D Convolutional Neural Network},
  journal = {Sensors},
  volume  = {19},
  number  = {11},
  pages   = {2472},
  year    = {2019},
  doi     = {10.3390/s19112472}
}

@inproceedings{hanson2019bidirectional,author    = {Hanson, Alex and PNVR, Koutilya and Krishnagopal, Sanjukta
               and Davis, Larry},
  title     = {Bidirectional Convolutional LSTM for the Detection of
               Violence in Videos},
  booktitle = {Computer Vision -- ECCV 2018 Workshops},
  series    = {Lecture Notes in Computer Science},
  volume    = {11130},
  pages     = {280--295},
  year      = {2019},
  publisher = {Springer},
  doi       = {10.1007/978-3-030-11012-3_24}
}

@article{chandra2026deep,author  = {Chandra, Rohitash and Li, Jiaqian and Dong, Ying and
             Zhuang, Henghao and Wu, Dongming},
  title   = {Deep Learning Framework for Video-Based Violence and
             Abuse Detection in Movies},
  journal = {Research Square},
  year    = {2026},
  note    = {Preprint},
  doi     = {10.21203/rs.3.rs-10061173/v1}
}

@inproceedings{valenzise2007scream,author    = {Valenzise, Giuseppe and Gerosa, Luigi and
               Tagliasacchi, Marco and Antonacci, Fabio and Sarti, Augusto},
  title     = {Scream and Gunshot Detection and Localization for
               Audio-Surveillance Systems},
  booktitle = {Proceedings of the 2007 IEEE Conference on Advanced
               Video and Signal Based Surveillance},
  pages     = {21--26},
  year      = {2007},
  publisher = {IEEE},
  doi       = {10.1109/AVSS.2007.4425280}
}

@inproceedings{vonboguszewski2021hateful,author    = {von Boguszewski, Niklas and Moin, Sana and
               Bhowmick, Anirban and Yimam, Seid Muhie and
               Biemann, Chris},
  title     = {How Hateful Are Movies? A Study and Prediction on
               Movie Subtitles},
  booktitle = {Proceedings of the 17th Conference on Natural
               Language Processing},
  pages     = {37--48},
  address   = {D{\"u}sseldorf, Germany},
  publisher = {KONVENS 2021 Organizers},
  year      = {2021}
}

@inproceedings{nam1998audiovisual,author    = {Nam, Jeho and Alghoniemy, Masoud and Tewfik, Ahmed H.},
  title     = {Audio-Visual Content-Based Violent Scene Characterization},
  booktitle = {Proceedings of the IEEE International Conference on Image Processing},
  volume    = {1},
  pages     = {353--357},
  year      = {1998},
  publisher = {IEEE}
}

@inproceedings{demarty2014benchmarking,author    = {Demarty, Claire-H{\'e}l{\`e}ne and Ionescu, Bogdan and
               Jiang, Yu-Gang and Lam, Vu and Schedl, Markus and Penet, C{\'e}dric},
  title     = {Benchmarking Violent Scenes Detection in Movies},
  booktitle = {Proceedings of the International Workshop on
               Content-Based Multimedia Indexing},
  pages     = {1--6},
  year      = {2014},
  publisher = {IEEE},
  doi       = {10.1109/CBMI.2014.6849827}
}

@article{pang2023audiovisual,author  = {Pang, Wenfeng and Xie, Wei and He, Qianhua and
             Li, Yanxiong and Yang, Jichen},
  title   = {Audiovisual Dependency Attention for Violence Detection
             in Videos},
  journal = {IEEE Transactions on Multimedia},
  volume  = {25},
  pages   = {4922--4932},
  year    = {2023},
  doi     = {10.1109/TMM.2022.3184533}
}

@article{xie2022optical,
  author  = {Xiao, Yang and Gao, Guxue and Wang, Liejun and Lai, Huicheng},
  title   = {Optical Flow-Aware-Based Multi-Modal Fusion Network for
             Violence Detection},
  journal = {Entropy},
  volume  = {24},
  number  = {7},
  pages   = {939},
  year    = {2022},
  doi     = {10.3390/e24070939},
  url     = {https://doi.org/10.3390/e24070939}
}

@article{pu2022semantic,author  = {Pu, Yujiang and Wu, Xiaoyu and Wang, Shengjin and
             Huang, Yuming and Liu, Zihao and Gu, Chaonan},
  title   = {Semantic Multimodal Violence Detection Based on
             Local-to-Global Embedding},
  journal = {Neurocomputing},
  volume  = {514},
  pages   = {148--161},
  year    = {2022},
  doi     = {10.1016/j.neucom.2022.09.090}
}

@article{peng2024hyperbolic,author  = {Peng, Xiaogang and Wen, Hao and Luo, Yikai and Zhou, Xiao and
             Yu, Keyang and Yang, Ping and Wu, Zizhao},
  title   = {Learning Weakly Supervised Audio-Visual Violence Detection
             in Hyperbolic Space},
  journal = {Image and Vision Computing},
  volume  = {151},
  pages   = {105286},
  year    = {2024},
  doi     = {10.1016/j.imavis.2024.105286}
}

@article{jin2025aligning,author  = {Jin, Wenping and Zhu, Li and Sun, Jing},
  title   = {Aligning First, Then Fusing: A Novel Weakly Supervised
             Multimodal Violence Detection Method},
  journal = {Knowledge-Based Systems},
  volume  = {322},
  pages   = {113709},
  year    = {2025},
  doi     = {10.1016/j.knosys.2025.113709}
}

@inproceedings{sultani2018real,author    = {Sultani, Waqas and Chen, Chen and Shah, Mubarak},
  title     = {Real-World Anomaly Detection in Surveillance Videos},
  booktitle = {Proceedings of the IEEE Conference on Computer Vision
               and Pattern Recognition},
  pages     = {6479--6488},
  year      = {2018}
}

@inproceedings{feng2021mist,author    = {Feng, Jia-Chang and Hong, Fa-Ting and Zheng, Wei-Shi},
  title     = {{MIST}: Multiple Instance Self-Training Framework for
               Video Anomaly Detection},
  booktitle = {Proceedings of the IEEE/CVF Conference on Computer
               Vision and Pattern Recognition},
  pages     = {14009--14018},
  year      = {2021}
}

@inproceedings{tian2021weakly,author    = {Tian, Yu and Pang, Guansong and Chen, Yuanhong and
               Singh, Rajvinder and Verjans, Johan W. and
               Carneiro, Gustavo},
  title     = {Weakly-Supervised Video Anomaly Detection with Robust
               Temporal Feature Magnitude Learning},
  booktitle = {Proceedings of the IEEE/CVF International Conference
               on Computer Vision},
  pages     = {4975--4986},
  year      = {2021}
}

@inproceedings{zhang2020span,author    = {Zhang, Hao and Sun, Aixin and Jing, Wei and
               Zhou, Joey Tianyi},
  title     = {Span-Based Localizing Network for Natural Language
               Video Localization},
  booktitle = {Proceedings of the 58th Annual Meeting of the
               Association for Computational Linguistics},
  pages     = {6543--6554},
  year      = {2020},
  publisher = {Association for Computational Linguistics},
  doi       = {10.18653/v1/2020.acl-main.585}
}

@inproceedings{lei2021detecting,author    = {Lei, Jie and Berg, Tamara L. and Bansal, Mohit},
  title     = {Detecting Moments and Highlights in Videos via
               Natural Language Queries},
  booktitle = {Advances in Neural Information Processing Systems},
  volume    = {34},
  pages     = {11846--11858},
  year      = {2021}
}

@inproceedings{yang2024text,author    = {Yang, Zhiwei and Liu, Jing and Wu, Peng},
  title     = {Text Prompt with Normality Guidance for Weakly
               Supervised Video Anomaly Detection},
  booktitle = {Proceedings of the IEEE/CVF Conference on Computer
               Vision and Pattern Recognition},
  pages     = {18899--18908},
  year      = {2024}
}

@article{bianculli2020dataset,title   = {A Dataset for Automatic Violence Detection in Videos},
  author  = {Bianculli, Miriana and Falcionelli, Nicola and Sernani, Paolo
             and Tomassini, Selene and Contardo, Paolo and Lombardi, Mara
             and Dragoni, Aldo Franco},
  journal = {Data in Brief},
  volume  = {33},
  pages   = {106587},
  year    = {2020},
  doi     = {10.1016/j.dib.2020.106587}
}

@inproceedings{radford2021clip,title={Learning Transferable Visual Models from Natural Language Supervision},
author={Radford, Alec and Kim, Jong Wook and Hallacy, Chris and Ramesh, Aditya and Goh, Gabriel and Agarwal, Sandhini and Sastry, Girish and Askell, Amanda and Mishkin, Pamela and Clark, Jack and Krueger, Gretchen and Sutskever, Ilya},
booktitle={Proceedings of the 38th International Conference on Machine Learning},
volume={139},
pages={8748--8763},
year={2021},
publisher={PMLR}
}

@inproceedings{chen2023beats,title={BEATs: Audio Pre-Training with Acoustic Tokenizers},
author={Chen, Sanyuan and Wu, Yu and Wang, Chengyi and Liu, Shujie and Tompkins, Daniel and Chen, Zhuo and Che, Wanxiang and Yu, Xiangzhan and Wei, Furu},
booktitle={Proceedings of the 40th International Conference on Machine Learning},
volume={202},
pages={5178--5193},
year={2023},
publisher={PMLR}
}

@inproceedings{vaswani2017attention,title={Attention Is All You Need},
author={Vaswani, Ashish and Shazeer, Noam and Parmar, Niki and Uszkoreit, Jakob and Jones, Llion and Gomez, Aidan N. and Kaiser, Lukasz and Polosukhin, Illia},
booktitle={Advances in Neural Information Processing Systems},
volume={30},
year={2017}
}

@inproceedings{rao2020local,author = {Rao, Anyi and Xu, Linning and Xiong, Yu and
            Xu, Guodong and Huang, Qingqiu and Zhou, Bolei and
            Lin, Dahua},
  title = {A Local-to-Global Approach to Multi-Modal Movie
           Scene Segmentation},
  booktitle = {Proceedings of the IEEE/CVF Conference on
               Computer Vision and Pattern Recognition},
  year = {2020},
  pages = {10146--10155},
  doi = {10.1109/CVPR42600.2020.01016}
}

@misc{bbfc_happyness,author = {{British Board of Film Classification}},
  title = {{The Pursuit of Happyness}: Classification and Content Advice},
  year = {2007},
  howpublished = {\url{https://www.bbfc.co.uk/release/the-pursuit-of-happyness-q29sbgvjdglvbjpwwc0zntm0ndu}},
  note = {Home-entertainment classification. Accessed 8 October 2026}
}

@misc{bamid_hiddenman,author = {{Bureau of Audiovisual and Music Industry Development}},
  title = {{Hidden Man}: Classification Record, Certificate No. 1071632},
  year = {2018},
  howpublished = {\url{https://cinema.bamid.gov.tw/search/result?CertificateNumber=1071632&CertificateType=F}},
  note = {Taiwan. Issued 6 November 2018 for a 134-minute festival
          version. Accessed 8 October 2026}
}

@misc{bbfc_dunkirk,author = {{British Board of Film Classification}},
  title = {{Dunkirk}: Classification and Content Advice},
  year = {2017},
  howpublished = {\url{https://www.bbfc.co.uk/release/dunkirk-q29sbgvjdglvbjpwwc0zotu0mzg}},
  note = {Home-entertainment classification. Accessed 8 October 2026}
}

@misc{bbfc_fallout,author = {{British Board of Film Classification}},
  title = {{Mission: Impossible -- Fallout}: Classification and Content Advice},
  year = {2019},
  howpublished = {\url{https://www.bbfc.co.uk/release/mission-impossible-fallout-q29sbgvjdglvbjpwwc00ote0nte}},
  note = {Home-entertainment classification. Accessed 8 October 2026}
}

@misc{bbfc_johnwick,author = {{British Board of Film Classification}},
  title = {{John Wick}: Classification and Content Advice},
  year = {2015},
  howpublished = {\url{https://www.bbfc.co.uk/release/john-wick-q29sbgvjdglvbjpwwc0znzuxotg}},
  note = {Home-entertainment classification. Accessed 8 October 2026}
}

@misc{bbfc_ipman,author = {{British Board of Film Classification}},
  title = {{Ip Man}: Classification and Content Advice},
  year = {2009},
  howpublished = {\url{https://www.bbfc.co.uk/release/ip-man-q29sbgvjdglvbjpwwc00mty4nzc}},
  note = {Home-entertainment classification. Accessed 8 October 2026}
}






\end{document}